\documentclass[journal]{IEEEtran}

\usepackage{multirow}
\usepackage{amsmath}  
\usepackage{amssymb}  
\usepackage{color,array}
\usepackage{xcolor}
\usepackage{graphicx}
\usepackage{setspace}  
\usepackage{makecell}
\usepackage{subcaption}
\DeclareCaptionLabelSeparator{periodspace}{.\quad}
\usepackage{graphics} 
\usepackage{epsfig} 
\usepackage{times} 
\usepackage{amsmath} 
\usepackage{amssymb}  
\usepackage[linesnumbered,ruled,vlined]{algorithm2e}

\usepackage{booktabs} 
\usepackage{textcomp, gensymb}
\usepackage{gensymb}  
\usepackage{bm}  
\usepackage{underscore}  

\usepackage{amsmath} 
\usepackage{amssymb}  

\usepackage{threeparttable}
\usepackage{enumerate}
\usepackage{bbm}

\usepackage{bbding} 
\usepackage{cite}
\usepackage[colorlinks=true, linkcolor=blue, citecolor=blue, urlcolor=blue]{hyperref}
\usepackage{hhline}
\usepackage{colortbl}

\author{Hanjing Ye\textsuperscript{+}$^{1}$, Kuanqi Cai\textsuperscript{+}$^{2}$, Yu Zhan$^{1}$, Bingyi Xia$^{3}$, Arash Ajoudani*$^{2}$, and Hong Zhang*$^{1}$ 
\thanks{\textsuperscript{+}Equal contributions, $^{*}$Equal supervision.}
\thanks{$^{1}$H. Ye, Y. Zhan and H. Zhang are with Robotics and Computer Vision (RCV) Laboratory, Southern University of Science and Technology (SUSTech), Shenzhen, 518000, China. $^{3}$B. Xia is with SUSTech.}%
\thanks{$^{2}$K. Cai and A. Ajoudani are with the Human-Robot Interfaces and Interaction (HRI$^2$) Laboratory, Istituto Italiano Di Tecnologia (IIT), Genova, 16163, Italy. K. Cai is also with the Swiss Federal Technology Institute of Lausanne. Correponding author: Hong Zhang (hzhang@sustech.edu.cn)}
}

\begin{document}
\title{\LARGE \bf
RPF-Search: Field-based Search for Robot Person Following \\in Unknown Dynamic Environments
}

\maketitle
\thispagestyle{empty}
\pagestyle{empty}

\begin{abstract}
Autonomous robot person-following (RPF) systems are crucial for personal assistance and security but suffer from target loss due to occlusions in dynamic, unknown environments. Current methods rely on pre-built maps and assume static environments, limiting their effectiveness in real-world settings. There is a critical gap in re-finding targets under topographic (e.g., walls, corners) and dynamic (e.g., moving pedestrians) occlusions.
In this paper, we propose a novel heuristic-guided search framework that dynamically builds environmental maps while following the target, and explicitly addresses these two types of occlusions through distinct mechanisms. For topographic occlusions, a belief-guided search field estimates the likelihood of the target’s presence and guides search toward promising frontiers. For dynamic occlusions, an observation-based search strategy adaptively switches between a fluid-following field and an overtaking potential field based on occluder motion patterns. Our results demonstrate that the proposed method outperforms existing approaches in terms of search efficiency and success rates, both in simulations and real-world tests. Our target search method enhances the adaptability and reliability of RPF systems in unknown and dynamic environments, supporting their use in real-world applications. Code, video, and experimental results are available at \href{https://medlartea.github.io/rpf-search/}{\textcolor{blue}{https://medlartea.github.io/rpf-search/}}.
\end{abstract}

\begin{figure*}[t]
    \centering
    \subfloat[Topographic occlusion]{\includegraphics[width=0.32\textwidth]{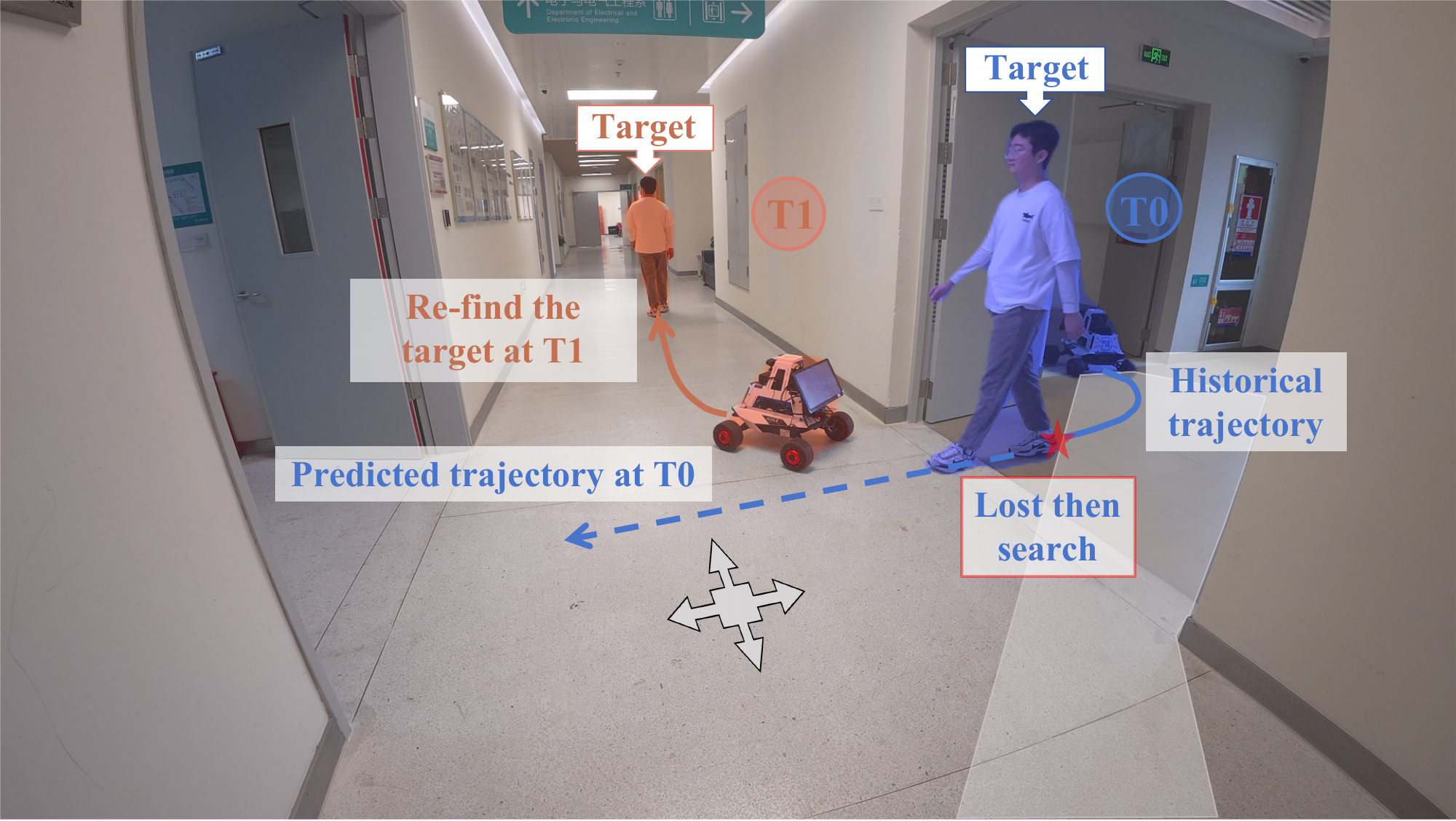}%
    \label{fig:coverTopographic}}
    \hspace{0.0005\linewidth}
    \subfloat[Dynamic occlusion with overtaking space]{\includegraphics[width=0.32\textwidth]{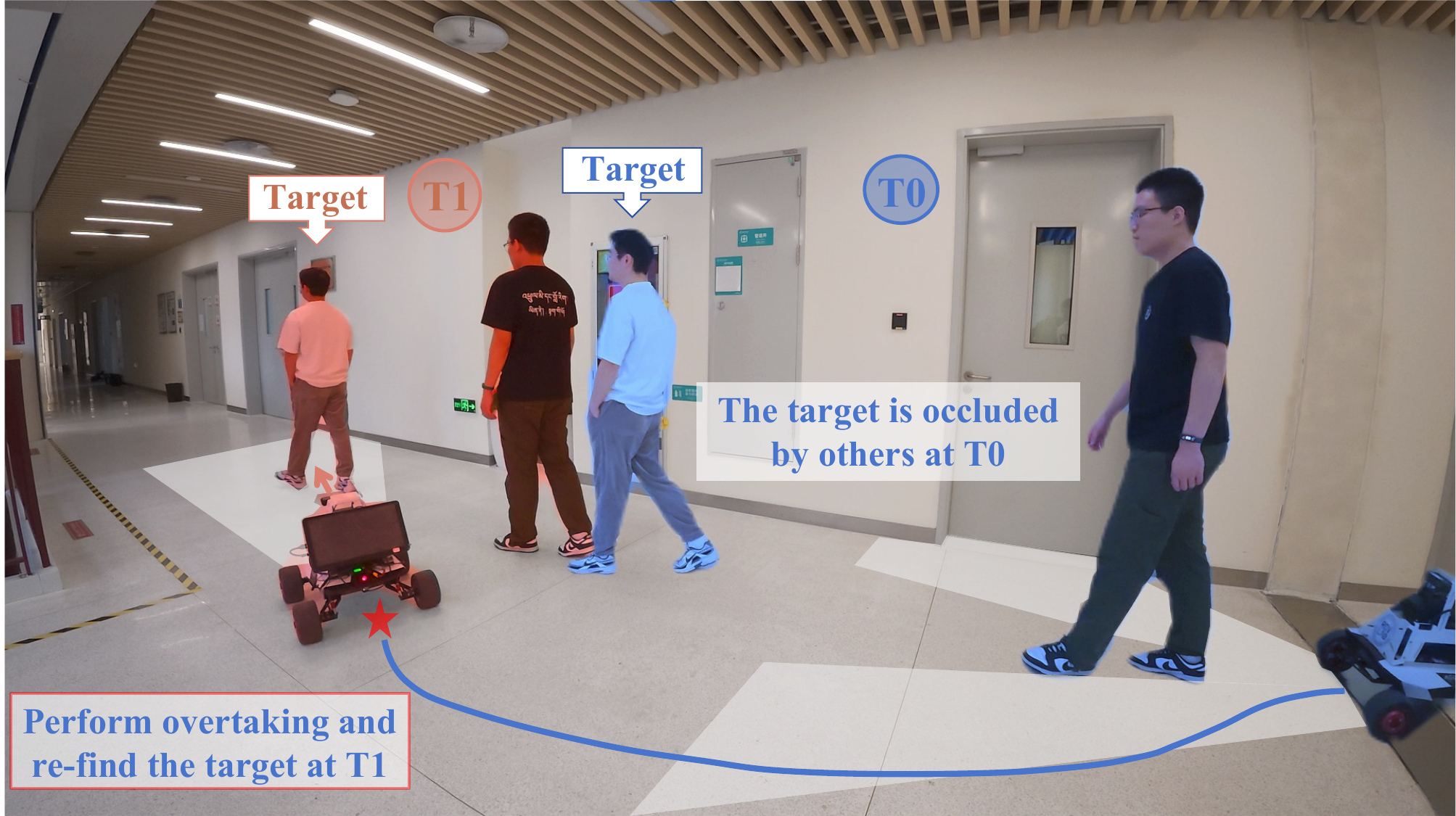}%
    \label{fig:coverSingle}}
    \hspace{0.0005\linewidth} 
    \subfloat[Dynamic occlusion without overtaking space]{\includegraphics[width=0.32\textwidth]{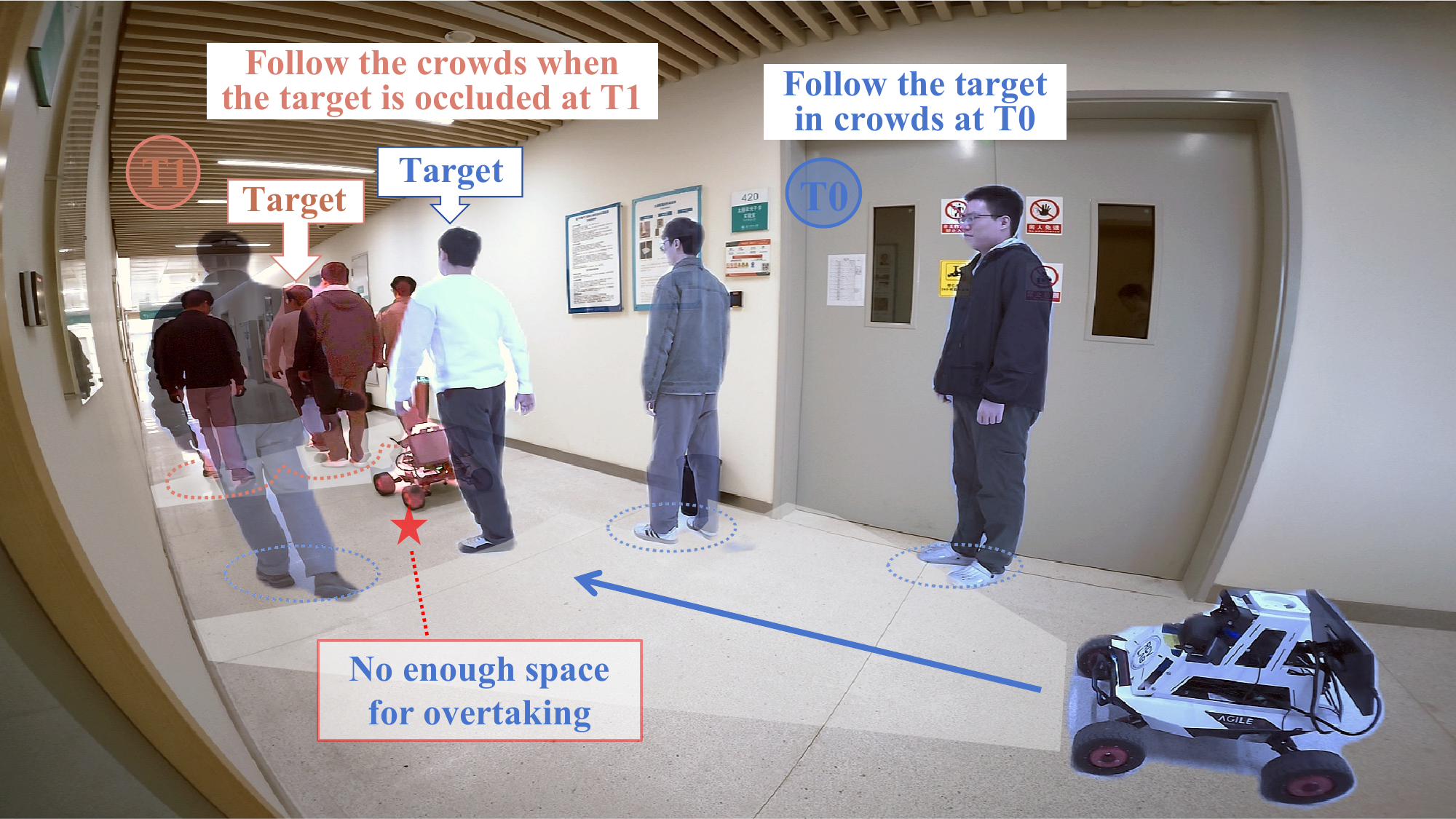}%
    \label{fig:coverMulti}}
\caption{RPF-Search addresses the challenges of person search in unknown, dynamic environments. The proposed approach enables the system to search for the target person in promising directions under topographic occlusion, as demonstrated in (a), despite a rough trajectory prediction. Additionally, the proposed method adapts to dynamic occlusions, either overtaking the occluders (b) or following them (c) based on their motion patterns and observed environments.}
\label{fig:cover}
\vspace{-15pt}
\end{figure*}


\section{INTRODUCTION}

\IEEEPARstart{R}{obot} person following\cite{yan2021human, chen2019human} (RPF) plays a vital role in human-robot interaction, with applications in personal assistance, security, and service robots. RPF is essential for tasks like elderly care and guided tours, especially in complex and dynamic environments.
However, target occlusions due to static objects or moving pedestrians can cause tracking failures.
Although active planning is a possible method for avoiding occlusion, it requires accurate trajectory predictions of humans and accurate obstacle detections for simulating the occlusion. These are difficult to obtain in realistic, unknown and dynamic environments. Furthermore, active occlusion avoidance demands sufficient free space for replanning, an assumption that often breaks down. For example, as shown in Fig.~\ref{fig:coverTopographic}, the doorway and hallway afford virtually no space for any re-planning maneuvers.
These point to a need for the robot to perform a search task to re-find the target person – a critical component of robust RPF systems, as illustrated in Fig.~\ref{fig:cover}.

Traditional search methods assume known environments and treat the task as a coverage problem\cite{isler2005tro}, maximizing visibility within limited timeframes. These approaches use prior information about both the environment and the target person, such as metric graphs\cite{goldhoorn2017search}, topo-metric graphs\cite{isler2005tro,Bayoumi2019search}, or semantic graphs\cite{Lilli2020ras} to locate lost targets. Subsequent searches rely on criteria like frequently visited nodes in topometric graphs or objects associated with the person in semantic graphs. While effective in static or known environments, these methods often struggle in complex and unknown settings. Exploration-based techniques\cite{basilico2011exploration,placed2023ral,niroui2019ral} select goals from frontiers of the searched region via utility metrics or deep reinforcement learning\cite{niroui2019ral}, but neglect motion cues of the lost person, resulting in inefficient searches.

Additionally, some methods leverage trajectory prediction models, such as linear regression\cite{algabri2021target}, Bayesian regression\cite{Lee2018icra}, and SVM-based regression\cite{Kim2018AnAF}, to estimate the target's future position. If unsuccessful, the search often defaults to frontier navigation, repeating the process\cite{Kim2018AnAF}. Such a greedy strategy ignores environmental dynamics, leading to extended searches. Furthermore, these methods neglect the dynamic behavior of other people in a real environment, resulting in dynamic occlusion of the target person and the risk of robot collision.

To address these limitations, we propose a field-based heuristic-guided search framework that incrementally maps the environment while following the target. The framework prioritizes areas with a high probability of target existence while maximizing environmental exploration and adapts to different types of occlusions:

\begin{itemize}
    \item \textbf{Topographic Occlusion}: For static obstacles like walls or corners, a belief-guided search field estimates target probability, optimizing information gain, trajectory prediction, path efficiency, and collision avoidance.
    
    \item \textbf{Dynamic Occlusion}: When moving pedestrians obstruct the target, the robot switches to a fluid-following field, to follow the occluders or adopts an observation-based potential field to overtake them by analyzing their motion patterns.
\end{itemize}

Unlike traditional map-dependent methods, our approach dynamically constructs maps during RPF by focusing on high-probability areas instead of uniform exploration typically employed in active SLAM \cite{basilico2011exploration, placed2023ral}. This probability representation is derived from our proposed incrementally constructed search fields, which leverage target motion cues and recently observed environmental information. As illustrated in Fig.~\ref{fig:framework}, this adaptive strategy ensures efficient and responsive searching in evolving environments, thereby enhancing the reliability of an RPF system.

\section{Related Work}

\subsection{Person Search Under Topographic Occlusion}

Person search in RPF aims to relocate a lost target. Existing methods often estimate the person’s position using a pre-built map. For example, Goldhoorn et al.\cite{goldhoorn2014continuous} employ a Partially Observable Markov Decision Process (POMDP) algorithm to estimate the target location on an occupancy map. To mitigate POMDP’s high computational cost, they later introduced a particle-filter-based approach\cite{goldhoorn2017search}, propagating particles from the target’s last known location and assigning higher weights to those in visible free spaces. The particle with the highest weight is selected as the next search goal.

Beyond occupancy maps, topo-metric graphs\cite{isler2005tro,Bayoumi2019search} and semantic graphs\cite{Lilli2020ras} have been used. For instance, Isler et al.\cite{isler2005tro} employ polygon triangulation to transform a metric map into a tree and use game theory to identify the optimal search node. Some methods incorporate prior knowledge to expedite the search. Bruckschen et al.\cite{Bayoumi2019search} utilize frequently visited locations in a topo-metric graph, modeling belief states with a hidden Markov model and updating them based on new observations. Similarly, semantic graphs have been used to predict navigation goals based on human-object interactions\cite{Lilli2020ras}. These approaches rely on prior information, often framing person search as a coverage problem. However, such prior knowledge is difficult to obtain in real-world RPF scenarios. A practical RPF system must enable robots to follow targets seamlessly in unfamiliar environments, requiring a way to search without pre-built maps.
Some methods bypass the need for prior information by leveraging historical trajectories to predict likely destinations. To handle sudden occlusions, techniques such as Kalman filters\cite{do2017reliable}, variational Bayesian regression\cite{Lee2018icra}, and least squares linear regression\cite{algabri2021target} forecast future trajectories. Although map-independent, these methods rely on single predictions, limiting their adaptability to handle unknown environments.

Person search in unknown environments inherently combines exploration with target search. Active SLAM\cite{placed2023ral} is a closely related research topic, focusing on efficient exploration by selecting optimal frontiers using utility-based metrics\cite{basilico2011exploration} or deep reinforcement learning policies\cite{niroui2019ral}. Placed et al.\cite{placed2023ral} introduce a graph-SLAM utility function to balance localization accuracy and exploration efficiency. Unlike these methods, which prioritize mapping, our approach focuses on effective person search. By detecting frontiers, we construct a belief-guided search field representing the likelihood of the target location. This field is incrementally updated using the target's past motion cues and recent environmental observations, ensuring efficient and adaptive search.

\subsection{Person Search Under Dynamic Occlusion}

Following a target person in a dynamic environment is challenging, as objects such as people can obscure a robot's view, causing tracking failures. Effective strategies for searching and re-identifying the target after occlusion are essential. A key difficulty lies in navigating through a dynamic crowd, where one approach is to follow the crowd until the target is re-identified. Crowd-aware navigation methods address this challenge by modeling pedestrian flows. Henry et al.\cite{henry} used inverse reinforcement learning (IRL) with Gaussian processes for human-like navigation. Cai et al.\cite{cai} proposed a sampling-based flow map algorithm but assumed complete knowledge of pedestrian positions. Dugas et al.\cite{Daniel} introduced a pseudo-fluid model with an observation-based flow-aware planner, while Jacques et al.\cite{saraydaryan2018navigation} developed the ``Flow Grid'' model, enabling the robot to adapt to human motion and optimize paths using a modified A* algorithm.
These methods excel in dense crowds or with fast-moving occluders but are less effective when occluders are sparse or slow. In such scenarios, overtaking occluders to re-identify the target becomes a practical and efficient strategy.

Overtaking maneuvers have been widely studied in autonomous driving \cite{lodhi2023autonomous}, with approaches generally divided into learning-based and model-based methods. Learning-based methods, such as Chen et al.\cite{chen2021automatic} applying deep reinforcement learning, model human-like overtaking but require large datasets and extensive simulations and face challenges in generalization and explainability.
In contrast, model-based methods offer greater interpretability and efficiency. Schwarting et al.\cite{schwarting2017parallel} developed a nonlinear model predictive control (NMPC) approach accounting for road boundaries, vehicle dynamics, and uncertainties in other traffic participants. Ma et al.\cite{ma2023overtaking} used potential field-based methods to enhance safety and comfort during overtaking by incorporating velocity, acceleration, and road boundaries. While effective in vehicular contexts, these methods are not designed for search tasks.

This paper presents a potential field-based strategy to handle occlusions in dynamic environments. By adaptively switching between following and overtaking occluders based on their motion cues, the proposed approach ensures efficient target re-identification in diverse social navigation scenarios.

\begin{figure*}[t]
    \centering
    \includegraphics[trim=0 0 0 0, clip, width=0.78\linewidth]{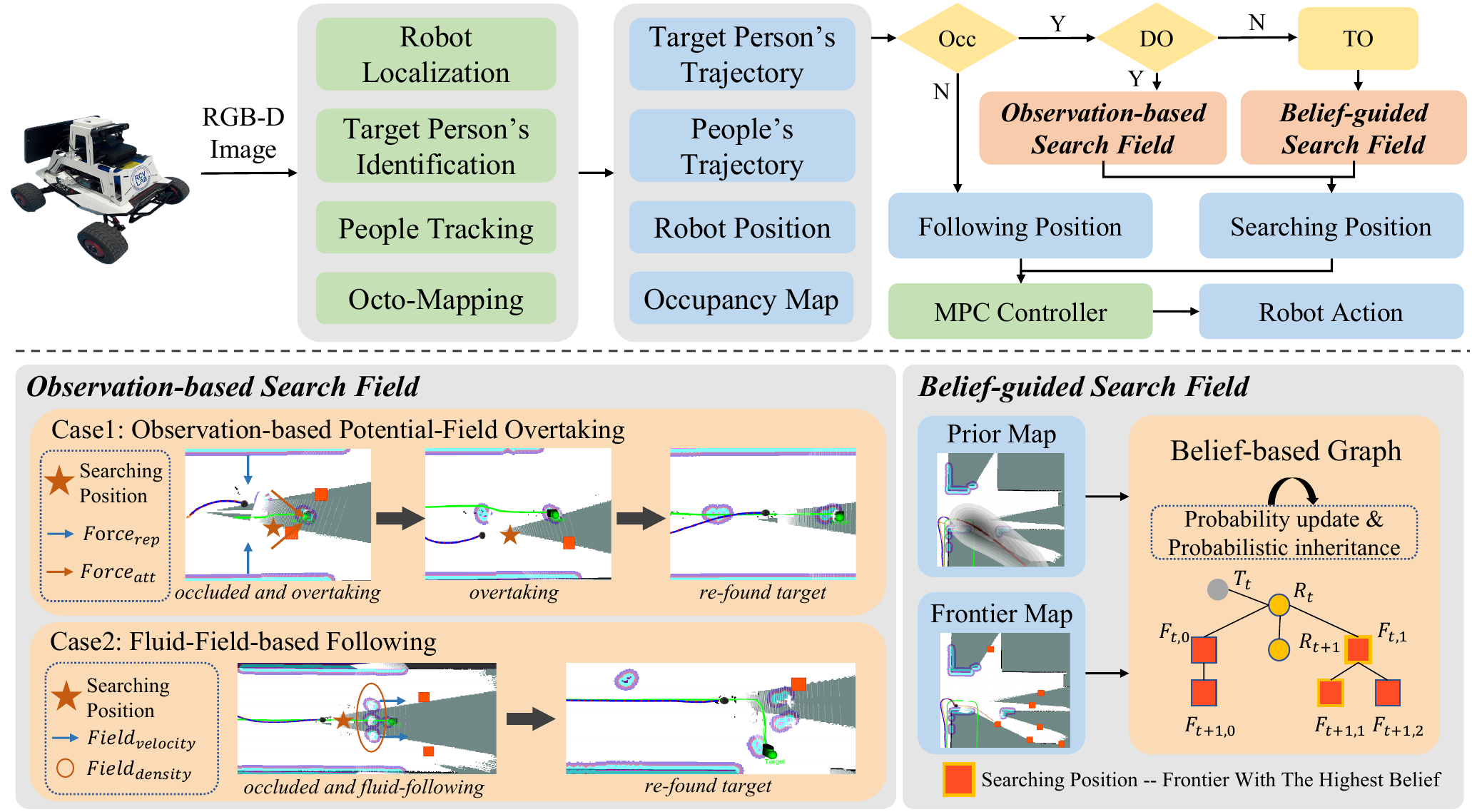}
    \caption{The framework of our RPF-search consists of three conditions: \textbf{Occ}, \textbf{DO}, and \textbf{TO}, representing ``is occlusion'', ``is dynamic occlusion'' and ``topographic occlusion'', respectively. The search process is divided into two categories: topographic occlusion from environmental corners and dynamic occlusion caused by moving pedestrians. For topographic occlusion, we use a belief-guided search strategy to optimize target search by selecting promising searching points based on information gain, target probability, motion efficiency, and collision risk. For dynamic occlusion, motion cues from pedestrians guide whether to follow via a fluid-field approach or overtake using an observation-based potential field.}
    \label{fig:framework}
\vspace{-15pt}
\end{figure*}

\begin{figure}
    \centering
    \includegraphics[trim=10 1 7 1, clip, width=0.8\linewidth]{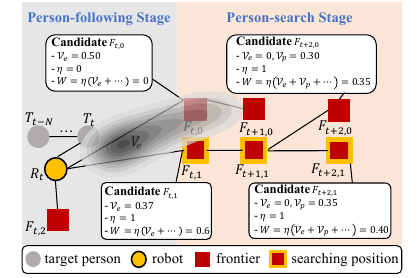}
    \caption{Simplified diagram of the belief-guided search field update process. When the target is lost, the existence probability $\mathcal{V}_e$, estimated through trajectory prediction, guides the search. The inference factor $\eta$ penalizes unlikely candidates based on path simulations. To handle limited visibility, probability inheritance $\mathcal{V}_p$ balances exploration inside and outside the room. During each timestep of the search stage, the frontier with the highest weight (maximum $W$) is selected as the next searching position.
}
    \label{fig:beliefFieldMethod}
\vspace{-15pt}
\end{figure}

\section{Method} \label{sec:Method}
The framework of the proposed method is illustrated in Fig.~\ref{fig:framework}. Our approach tracks people and identifies the target person based on our previous work \cite{ye2023reid}. Additionally, we utilize visual-inertial odometry for robot localization and employ Octomap \cite{hornung13auro} to map unknown environments. Octomap is a probabilistic-based mapping method that remains robust to dynamic objects. If no occlusion occurs, the robot navigates to a predefined person-following position—typically a fixed location behind the target—using a model predictive controller (MPC)\cite{zhao2024human}, enabling simultaneous person-following and obstacle avoidance. In the event of occlusion, the robot instead leverages the aforementioned perception information to locate and navigate to the best position for re-finding the target person in unknown and dynamic environments. 

To differentiate between types of occlusions—topographic and dynamic—we use an intersection-over-union (IoU) analysis. If the occluder is identified as a pedestrian via the IoU assessment (i.e., the target person’s bounding box previously interacted with another pedestrian), the system activates an \textbf{Observation-based Search Field} (Sec.~\ref{sec:DynaSearch}) to resume target tracking under dynamic occlusions. In contrast, the presence of static obstacles prompts the deployment of a \textbf{Belief-guided Search Field} (Sec.~\ref{sec:TopoSearch}), enabling smooth adaptation to static occlusions. Detailed method flow can be found in the supplementary materials.

\subsection{Handling topographic occlusions: Belief-guided search field} \label{sec:TopoSearch}
Our approach identifies the optimal search goal using the partially built map, past target trajectory data, and the robot's current position. It involves three steps: candidate generation, weight calculation, and belief-guided field establishment with a simplified diagram shown in Fig.~\ref{fig:beliefFieldMethod}.

\subsubsection{Candidate Generation}
When the target disappears, the robot constructs a belief-guided field to represent the probability of the target's location. Search candidates are generated using current and historical observation to focus on high-belief regions. In a partially mapped environment, a key issue in our solution involves identifying \emph{frontiers}, points on the boundary between mapped and unknown regions. These frontiers serve as both observation points and safe navigation locations due to their proximity to mapped free spaces. The belief-guided field is initialized with high values at these frontier points.

Frontiers are detected using Canny Edge Detection on the map\cite{placed2023ral}, with contour centers identified as candidates  $\{\mathcal{F}_i\}$. These candidates form the high-belief regions guiding the robot's search. 
With these candidates, we define the belief-guided field and obtain the searching point $\mathbf{x}^*$ as follows:
\begin{equation}
    {\mathbf{x}^{*}} = \arg \max_{\mathbf{x}} \left\{ W_i \exp\left( -\frac{\| \mathbf{x} - \mathcal{F}_i \|^2}{2 \sigma^2} \right) \right\},
\label{eq:beliefField}
\end{equation}
where $\mathbf{x}=[x,y]$ represents a position in the environment, $\sigma$ is the Gaussian variance, and $W_i$ is the weight of the $i$-th candidate, estimated from the target's motion cues and environmental information. The maximum operator ensures that overlapping Gaussians prioritize the highest probability, capturing the most likely location.

\subsubsection{Candidate Weight Calculation}
The utility of each candidate is quantified by expected uncertainty in three target variables: information gain $\mathcal{V}_g$, collision risk $\mathcal{V}_c$, and target existence $\mathcal{V}_e$. The candidate weight is computed as:
\begin{equation}
    W_i = \eta_{i} \left( \mathcal{V}_g(\mathcal{F}_i) + \mathcal{V}_c(\mathcal{F}_i) + \mathcal{V}_e(\mathcal{F}_i) + \mathcal{V}_p(\mathcal{F}_i) \right),
\label{eq:objectiveFunction}
\end{equation}
where $\mathcal{V}_p$ represents probabilistic inheritance, capturing probability transfer from farther candidates, and $\mathcal{V}_g$, $\mathcal{V}_c$, and $\mathcal{V}_e$ are normalized terms explained below.

\textbf{Inference Factor} $\eta_{i}$: To reduce computational load and quickly locate the target, the inference factor \(\eta_i\) filters candidates based on the maximum distance a pedestrian could travel since last observed, calculated using the maximum velocity of pedestrian. If the distance between a candidate and the last known position of the target exceeds this estimate, it implies the candidate would have been observed if the target were moving in that direction. Thus, further weight calculation is unnecessary for such candidates. The inference factor \(\eta_i\) is defined as:
\begin{equation}
    \eta_{i} := \mathbb{I}\left( \Delta t \cdot v_{\max} > D_{\text{ijk}}(P_{\text{lost}}, \mathcal{F}_i) \right),
\label{eq:inferenceFactor}
\end{equation}
where $\mathbb{I}$ is the indicator function, returning one if the condition holds and zero otherwise. $\Delta t$ is the time elapsed since the target was last seen, and $D_{\text{ijk}}(P_{\text{lost}}, \mathcal{F}_i)$ is the Dijkstra-computed distance between the lost position $P_{\text{lost}}$ and the candidate $\mathcal{F}_i$ in the map.

\textbf{Information Gain} $\mathcal{V}_g(\cdot)$: Information gain prioritizes maximizing observation of unknown areas using \textit{entropy} to quantify exploration uncertainty~\cite{basilico2011exploration}. The expected entropy of a candidate is:
\begin{equation}
    \mathcal{V}_g(\mathcal{F}_i) = - \sum_{{\rm{c}} \in {\rm Adj}(\mathcal{F}_i)} p(c)\log_2(p(c)),
\label{eq:informationGain}
\end{equation}
where $p(c)$ is the occupancy probability of cell $c$, and ${\rm Adj}(\cdot)$ represents the adjacent cells of the candidate in the map.

\textbf{Collision Risk} $\mathcal{V}_c(\cdot)$: This factor prevents candidates from being too close to obstacles or pedestrians (non-target), reducing collision risk and avoiding invading human comfort space:
\begin{equation}
    \mathcal{V}_c(\mathcal{F}_i) = \frac{\max({\Phi}(\mathcal{F}_i))}{R(\mathcal{F}_i) + \delta},
\label{eq:collisionRisk}
\end{equation}
where ${\Phi}(\cdot)$ represents the comfortable region around humans~\cite{cai2022human}, $R(\cdot)$ is the collision probability with obstacles~\cite{fulgenzi2010risk}, and $\delta$ ensures numerical stability.

\textbf{Target Person Existence} ${\mathcal{V}_e}(\cdot)$: To efficiently locate the target, ${\mathcal{V}_e}$ evaluates candidates based on predicted potential locations when the target abruptly disappears. Unlike traditional trajectory prediction relying on global maps, ${\mathcal{V}_e}$ uses human motion cues to estimate potential directions within a limited observation map. It consists of two steps: 1) initial trajectory prediction and 2) probability propagation.

\paragraph{Initial Trajectory Prediction} 
A regression model is constructed based on the target’s historical motion cues, including velocity and positions, to forecast potential trajectories. We use support vector machine regression (SVR), which effectively captures nonlinear relationships and provides accurate fits in constrained environments~\cite{Kim2018AnAF}. The prediction is framed as finding a function, $[x_t, y_t]^T = \mathbf{w}^T\psi(t) + \mathbf{b}$. $\mathbf{w}$ is the weight vector, $\mathbf{b}$ is a constant bias vector, and $\psi$ represents a non-linear mapping. Using historical timestamps and positions of the target as the training dataset, the function is learned with SVR theory, enabling estimation of the target's position at any future timestamp.

\paragraph{Probability Propagation} 
Unlike trajectory predictions using accurate global maps, our algorithm predicts trajectories based on historical human cues and a limited map generated by the robot's observations. The predicted trajectory is transformed into a probabilistic representation using a Gaussian-based method combined with motion cues. The probability of a point in the trajectory is defined as:
\begin{equation}
    \mathcal{N}_{\mu(t), \Sigma(t)}(\mathbf{x}) = e^{-\frac{1}{2} (\mathbf{x}-\mu(t))^T\Sigma(t)^{-1}(\mathbf{x}-\mu(t))},
\end{equation}
where $\mu(t) = [\mu_{x,t}, \mu_{y,t}]$ represents a predicted point, and $\Sigma(t)$, the covariance matrix, reflects the probable area of the target based on velocity $v$ and social distance $d_s$~\cite{cai2022human}:
\begin{equation}
    \Sigma(t) = \mathrm{Rot}(\varphi_t) \cdot \mathrm{diag}(vd_s, d_s)^2 \cdot \mathrm{Rot}(\varphi_t)^T,
\end{equation}
where $\varphi_t$, estimated from adjacent points, defines the orientation, resulting in an elliptical probability area in the predicted direction.
For $N$ predicted items, the probability of the target's appearance is:
\begin{equation}
    \mathcal{V}_e(\mathbf{x}) = \sum_{t=t_0}^{t_N} \omega_t \mathcal{N}_{\mu(t), \Sigma(t)}(\mathbf{x}),
    \label{eq:targetPersonExistence}
\end{equation}
where $\omega_t$ is the weight of each component. To account for increasing uncertainty over time, weights decrease monotonically:
\begin{equation}
    \omega_t = \frac{e^{-\lambda t}}{\sum_{t=1}^{N} e^{-\lambda t}}, \quad \text{with} \quad \sum_{t=t_0}^{t_N} \omega_t = 1.
\end{equation}

\textbf{Probabilistic Inheritance} $\mathcal{V}_p(\cdot)$: 
In unstructured environments, obstacles can significantly restrict visibility, causing the robot to miss detections even at optimal candidate positions. For instance, in Room 1 (Fig.~\ref{fig:factory}), the target (blue dot) is hidden behind Obstacle 1 and Obstacle 2. Without a comprehensive exploration of the room, the robot is likely to
miss detecting the target person at this location. To address this, we propose a probabilistic inheritance strategy to balance exhaustive exploration of the room and outside the room.
The probability of a candidate $\mathcal{F}_i$ inherits from its parent candidate, $Parent(\mathcal{F}_i)$, the closest candidate from the previous generation. The inherited probability is defined as:
\begin{equation}
    \mathcal{V}_p(\mathcal{F}_i) = P_{\rm parent} \cdot e^{-d_p / \varepsilon},
\label{eq:probabilityInheritance}
\end{equation}
where $d_p$ is the distance between $\mathcal{F}_i$ and $Parent(\mathcal{F}_i)$, and the inheritance probability decays with increasing distance. Adjacent frame frontiers maintain spatial coherence in the environment, thereby enabling effective probability inheritance. The normalization factor $\varepsilon$ is adaptive to the room size, ensuring probability decay matches the environment. If candidates inside the room have lower probabilities than previous candidates, the robot transitions from searching inside the room to exploring outside. 

\subsection{Handling dynamic occlusions: Observation-based search field}  \label{sec:DynaSearch}
Another significant challenge in accurately tracking a target in dynamic environments is dynamic occlusion. Moving pedestrians can obstruct the robot’s view, causing it to lose track of the target. For instance, the target may enter a crowd or random people may walk between the robot and the target. Unlike the belief-guided field used for topographic occlusions, we propose an observation-based search field leveraging the motion cues of occluders to handle dynamic occlusions.

Dynamic occlusions are categorized into two types. \textbf{Temporary occlusion} occurs when a pedestrian briefly crosses between the target and the robot. Here, the occlusion duration is short, allowing the robot to quickly resume tracking without altering its strategy and maintaining the previous track. \textbf{Long-term occlusion} happens when occluders persist in the robot’s view, easily resulting in a track loss. In this case, we categorize occlusions into two types based on the motion cues of the occluder(s) and introduce two corresponding strategies: 1) \textit{Observation-Based Potential Field Overtake Strategy}, and  
2) \textit{Fluid Field-Based Following Strategy}.  
Both strategies use vector fields and are triggered when the target is obscured by occluders.

\begin{figure}[t]
    \centering
    \includegraphics[width=1.0\linewidth]{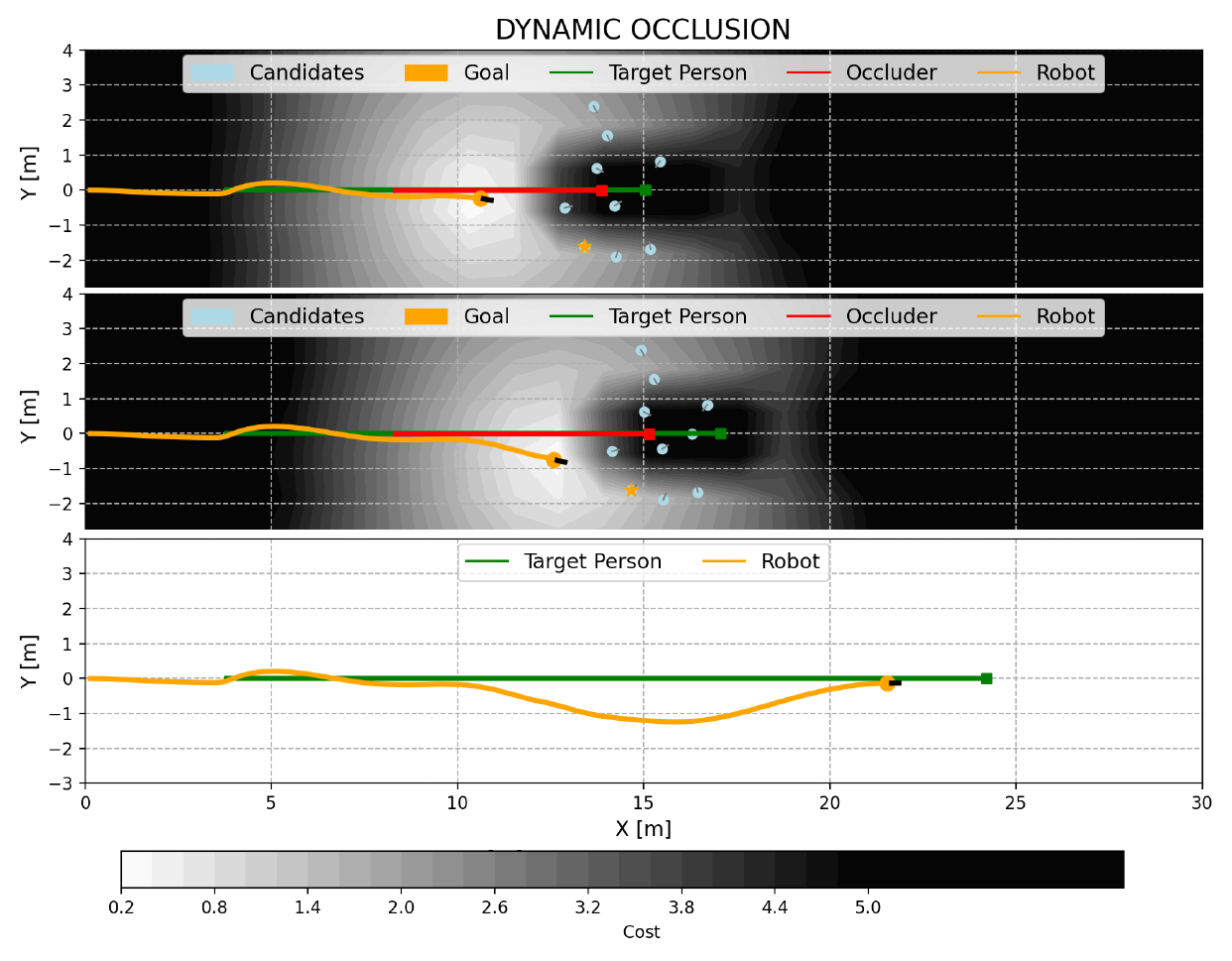}
    \caption{Overtaking Visualization: During dynamic occlusion, the robot constructs an overtaking field using the occupancy map and occluder motion cues to optimize target observation, path cost, and collision avoidance. Lighter areas indicate lower costs (max cost = 5.0). The robot overtakes the occluder (red) by selecting an optimal goal (orange) from candidate points (blue). Initially, the field is vertically symmetric (first row). As the robot moves to the lower region, reduced path costs in this area (lighter colors) lead the robot to consistently choose the same overtaking point, resulting in a smooth overtaking trajectory (orange path in the last row).}
    \label{fig:overTakeVis}
\vspace{-13pt}
\end{figure}

\begin{figure}[t]
    \centering
    \includegraphics[width=1.0\linewidth]{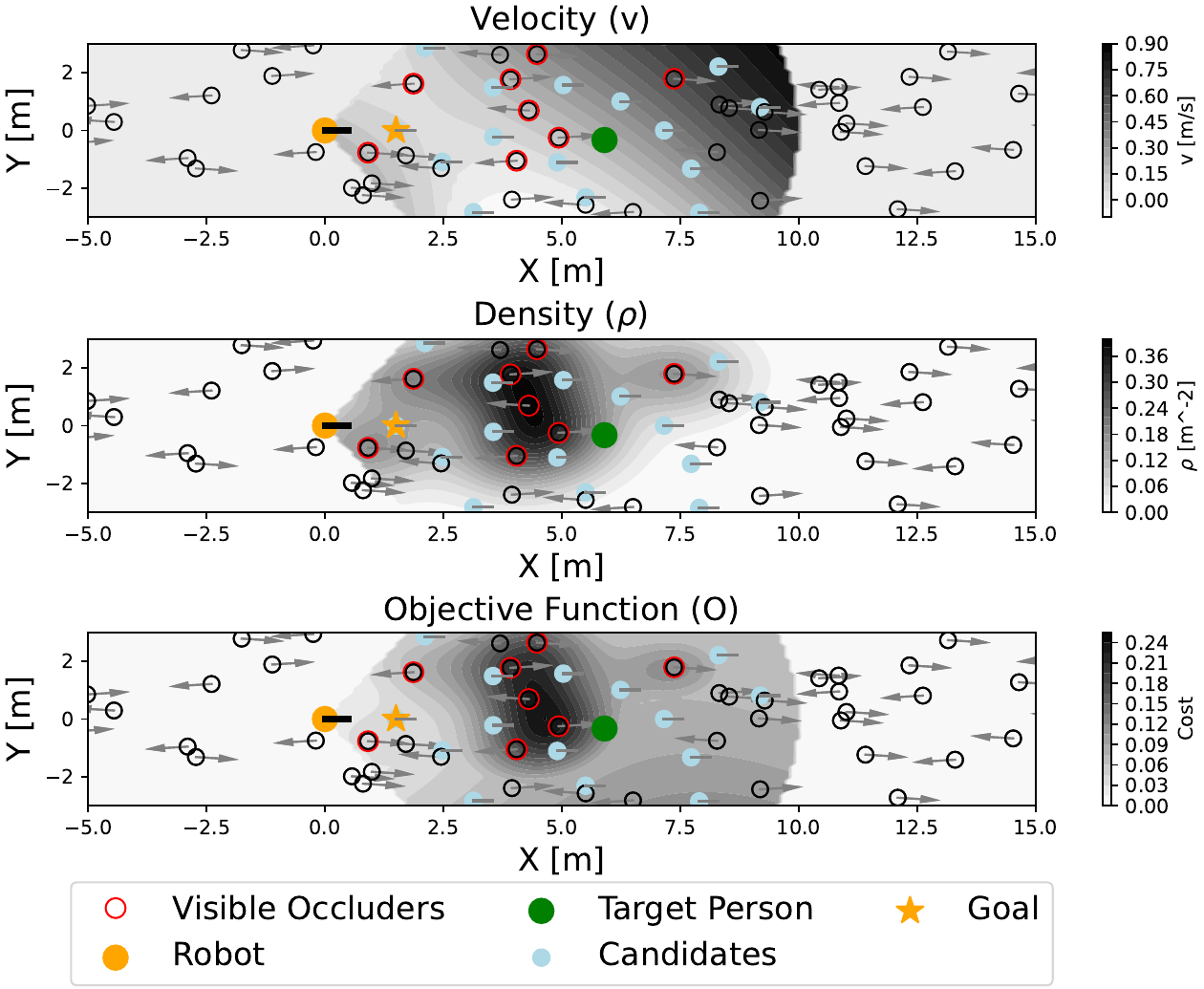}
    \caption{Fluid-Following Visualization: In multi-person occlusion scenarios where overtaking isn't feasible, the robot constructs a fluid-following field using motion cues from visible occluders (red circles). The first row displays the velocity field of the pseudo-fluid, the second row shows the density field to avoid high-density areas, and the final row combines these into the fluid-following field. Using this field, the robot selects an optimal goal (orange) from candidate points (blue), ensuring safe navigation along the pseudo-fluid path while avoiding high-density zones.}
    \label{fig:fluidFollowingVis}
\vspace{-13pt}
\end{figure}

\subsubsection{\textit{Observation-based potential field overtake strategy}}
When the target is lost due to dynamic occlusion, a natural human response is to overtake the occluders to regain visibility and resume following. To do this, we propose an observation-based potential field overtaking strategy leveraging occluder motion cues. The strategy predicts the occluder trajectory, $U_{\rm occluder}$, and constructs an overtaking field using repulsive and attractive forces inspired by the artificial potential field method.
The \textbf{repulsive force}~\cite{CaiIROS2024} prevents collisions with pedestrians (including occluders) and obstacles:
\begin{equation}
\mathbf{F}_{\text{rep}}(\mathbf{x}) = 
\begin{cases}
    \eta_f \left( \frac{1}{\|\varpi\|} - \frac{1}{d_0} \right) \frac{\varpi}{\|\varpi\|^3}, & \text{if } \|\varpi\| \leq d_0, \\
    0, & \text{if } \|\varpi\| > d_0,
\end{cases}
\end{equation}
\begin{equation}
\varpi = \mathbf{x} - \underset{\mathbf{u} \in U_{\rm occluder}}{\arg\min} \|\mathbf{x} - \mathbf{u}\|,
\end{equation}
where $\eta_f$ is the repulsive gain coefficient, and $d_0$ is the influence distance beyond which the repulsive force ceases.
The \textbf{attractive force} guides the robot to optimal positions, using a two-step process: \textbf{Filter} and \textbf{Plan}:
\begin{equation}
\mathbf{F}_{\text{att}}(\mathbf{x}) = 
\begin{cases}
    \eta_a (\mathbf{x}_{\text{robot}} - \mathbf{x}) / (G(\mathbf{x}|\mathbf{x}_{\rm occ},\bar{\mathbf{x}}_{\rm tar})+1), & \text{Filter}, \\
    \eta_a (\mathbf{x} - \mathbf{x}^*) / (G(\mathbf{x}|\mathbf{x}_{\rm occ},\bar{\mathbf{x}}_{\rm tar})+1), & \text{Plan},
\end{cases}
\end{equation}
where $\eta_a$ is the attractive gain coefficient, $\mathbf{x}_{\text{robot}}$ is the robot's current position, and $G(\mathbf{x}|\mathbf{x}_{\rm occ}, \bar{\mathbf{x}}_{\rm tar})$ represents the observation gain of a candidate $\mathbf{x}$ given the occluder position $\mathbf{x}_{\rm occ}$ and the predicted target position $\bar{\mathbf{x}}_{\rm tar}$. Here, $\bar{\mathbf{x}}_{\rm tar}$ is a predicted position in front of the occluder. Candidates with higher gains minimize occluder overlap in the image space from the robot’s camera perspective, facilitating better observation of $\bar{\mathbf{x}}_{\rm tar}$.

We follow the assumption from prior works on person tracking~\cite{koide2016fusion, ye2023track, ye2023reid}, where a person is projected into the image as a bounding box based on the camera's pose and projection function, $\text{Proj}(\cdot, \mathbf{x})$. The observation gain is defined as
$G(\mathbf{x}|\mathbf{x}_{\text{occ}},\bar{\mathbf{x}}_{\rm tar}) = 1 - \text{IoU}(\text{Proj}(\bar{\mathbf{x}}_{\rm tar}, \mathbf{x}), \text{Proj}(\mathbf{x}_{\text{occ}}, \mathbf{x})),$
where $\text{IoU}(\cdot, \cdot)$ is the intersection-over-union of two bounding boxes. Additionally, the attractive field incorporates path cost via the direction $\mathbf{x}_{\text{robot}} - \mathbf{x}$. The combined field is shown in Fig.~\ref{fig:overTakeVis}.

In the \textbf{filter step}, the observation-based potential field filters sampling points to identify the optimal candidate $\mathbf{x}^*$, which maximizes target observation and minimizes trajectory cost. To accelerate selection and ensure smooth navigation, the search area is restricted to the robot's front, with candidates $\{\mathbf{x}_i\}$ sampled using the Korobov lattice~\cite{korobov1963number}, a method for generating low-discrepancy points. The optimal candidate $\mathbf{x}^*$ is determined as:
\begin{equation}
\mathbf{x}^* = \mathop{\arg\min}_{\{\mathbf{x}_i\}} \|\mathbf{F}_{\text{att}}(\mathbf{x}_i)\| + \|\mathbf{F}_{\text{rep}}(\mathbf{x}_i)\|.
\end{equation}
In addition, we record its minimum cost as $C^*$. If $C^* > \delta_c$ and the occluders' maximum velocity is below $\delta_{v_{\text{min}}}$, the strategy switches to the belief-guided search field (Sec.~\ref{sec:TopoSearch}), suitable for global-scale planning in low-velocity, high-risk environments. If $C^* > \delta_c$ or the occluders' minimum velocity exceeds $\delta_{v_{\text{max}}}$, the strategy switches to the fluid-field-based following strategy (Sec.~\ref{followingStrategy}). Otherwise, $\mathbf{x}^*$ becomes the searching point. In the \textbf{plan} step, the attractive and repulsive fields are overlaid, and the gradient of the combined field determines the robot's movement direction to $\mathbf{x}^*$.

An example of the overtaking process is shown in Fig.~\ref{fig:overTakeVis}. A slowly moving distractor (red square) occludes the robot’s view, causing a target loss. In response, an overtaking field is generated, where lighter colors indicate lower costs. With sufficiently low overtaking cost, the robot selects the optimal goal (orange) from candidates (blue) and overtakes the occluder. Initially, the field is nearly symmetric (first row), with low-cost regions on both sides. As the robot moves to the lower region, the path-cost effect reduces the cost there (darker shade), leading to consistent selection of the same overtaking point and producing a smooth overtaking trajectory (orange path, last row).

\subsubsection{Fluid field-based following strategy} \label{followingStrategy}
To ensure natural and safe navigation in dynamic occlusion scenarios, following the occluding person is a viable strategy when overtaking is infeasible, such as when occluders move too fast or overtaking costs are too high, risking unsafe behavior. 
The goal of this strategy is for the robot to follow occluders effectively. Because moving too slowly increases the distance to the occluder, risking target loss, while overtaking may require acceleration, posing collision risks in human-shared environments.

Inspired by the fluid-based navigation model~\cite{DugasFlowBot,cai}, we propose treating detected pedestrians, including occluders, as a pseudo-fluid. This enables the estimation of a velocity field around the robot's perception, leading to safe and efficient target tracking. The fluid-based vector field is modeled using the following variables.

\textbf{Velocity Field} $\mathrm{v}(\mathbf{x}) = (\hat{\mathrm{v}}_x(\mathbf{x}), \hat{\mathrm{v}}_y(\mathbf{x}))$: To estimate the velocity at $\mathbf{x}$, we consider detected pedestrians $Z_p = \{(\mathbf{x}_j, \mathrm{v}_j)\}$ and use:
\begin{equation}
\hat{\mathrm{v}}_x(\mathbf{x}) = \frac{\sum_{j \in Z_p} \alpha \mathrm{v}_{x,j}}{\sum_{j \in Z_p} \alpha}, \quad
\hat{\mathrm{v}}_y(\mathbf{x}) = \frac{\sum_{j \in Z_p} \alpha \mathrm{v}_{y,j}}{\sum_{j \in Z_p} \alpha},
\end{equation}
where $\alpha = \exp(-\gamma \|\mathbf{x} - \mathbf{x}_j \|^2)$ and $\gamma$ controls the "viscosity" of the fluid flow~\cite{DugasFlowBot}.

\textbf{Density Field} $\rho(\mathbf{x})$: To avoid the robot rushing into high-density areas, the density field is defined as:
\begin{equation}
\rho(\mathbf{x}) = \frac{1}{2\pi\sigma^2} \frac{1}{N_{\text{obs}}(\mathbf{x})} \sum_{j \in Z_p} \exp\left( -\frac{\|\mathbf{x} - \mathbf{x}_j\|^2}{2\sigma^2} \right),
\end{equation}
where $N_{\text{obs}}(\mathbf{x})$ is the number of times $\mathbf{x}$ has been observed, encouraging navigation to familiar areas, and $\sigma$ is a constant to control density smoothness.

\textbf{Objective Function} $\mathcal{O}(\mathbf{x})$: The searching point $\mathbf{x}^*$ is obtained by minimizing:
\begin{equation}
\mathbf{x}^* = \mathop{\arg\min}_{\{\mathbf{x}_i\}} \gamma \rho(\mathbf{x}) \| \mathrm{v}_{\text{robot}} - \mathrm{v}(\mathbf{x}) \|,
\label{eq:fluidFollowingOptimization}
\end{equation}
where $\mathrm{v}_{\text{robot}}$ is the robot's velocity. Candidate points $\{\mathbf{x}_i\}$ are sampled in front of the robot, and the optimal candidate is selected as the searhcing point. The purpose of this objective function (Eq.~\ref{eq:fluidFollowingOptimization}) is to guide the robot in following occluders by minimizing the deviation from their velocity while accounting for their density. Fig.~\ref{fig:fluidFollowingVis} illustrates the process. With the proposed overtaking and fluid-following fields, our method effectively searches for the target person under dynamic occlusions, ensuring safe navigation in complex environments.

\section{Experiments}
This section evaluates the effectiveness of the proposed person-search method, particularly with regard to its performance in re-finding a lost target person due to topographic or dynamic occlusions, by comparing it with existing baseline methods under challenging conditions. The parameters of the proposed method are shown in the supplementary materials. 

\subsection{Experimental Settings}
\begin{figure}[t]
    \centering
    \begin{subfigure}[t]{0.30\textwidth}
        \centering
        \includegraphics[width=\textwidth]{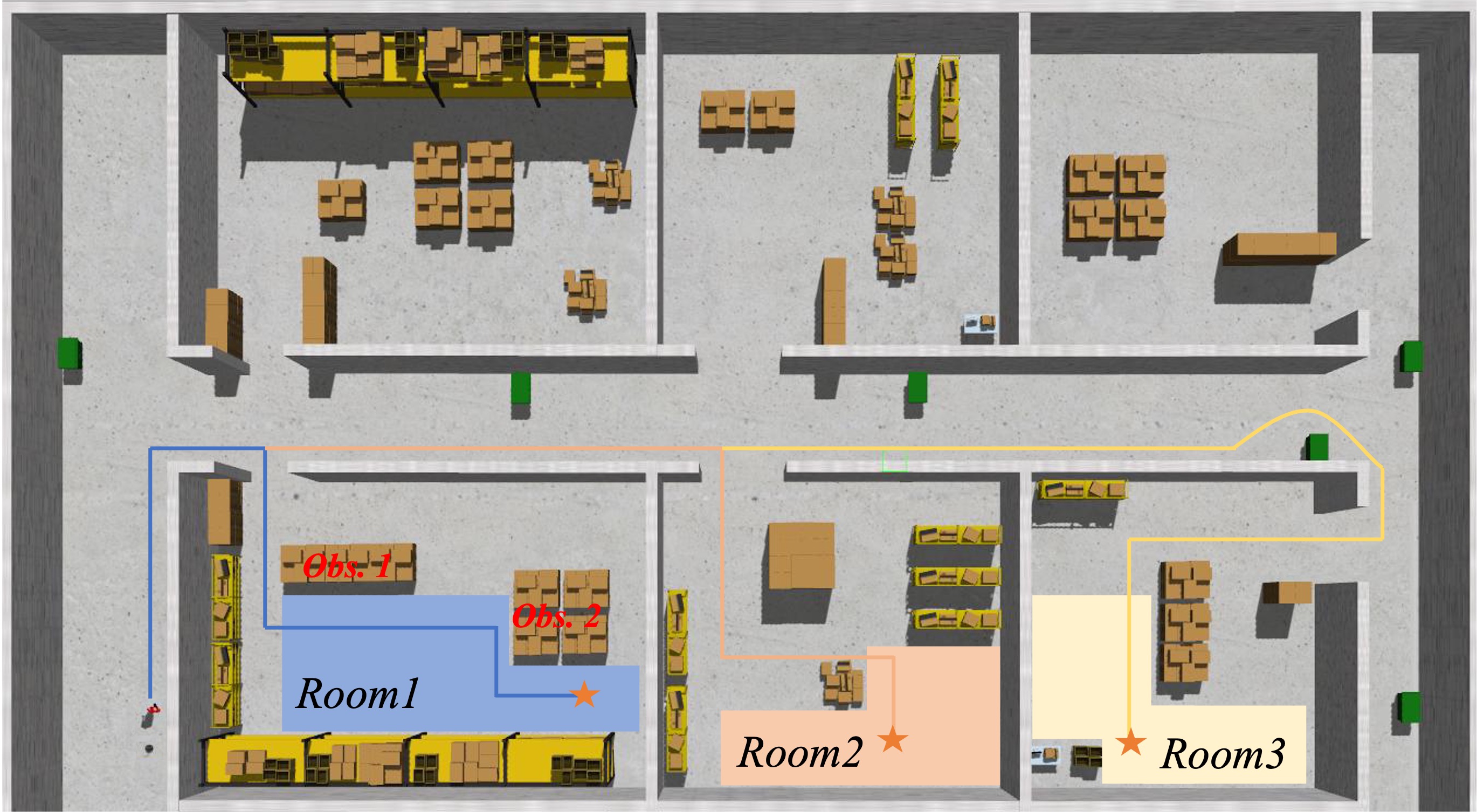}
        \caption{Factory Env.}
        \label{fig:factory}
    \end{subfigure}%
    \begin{subfigure}[t]{0.172\textwidth}
        \centering
        \includegraphics[width=\textwidth]{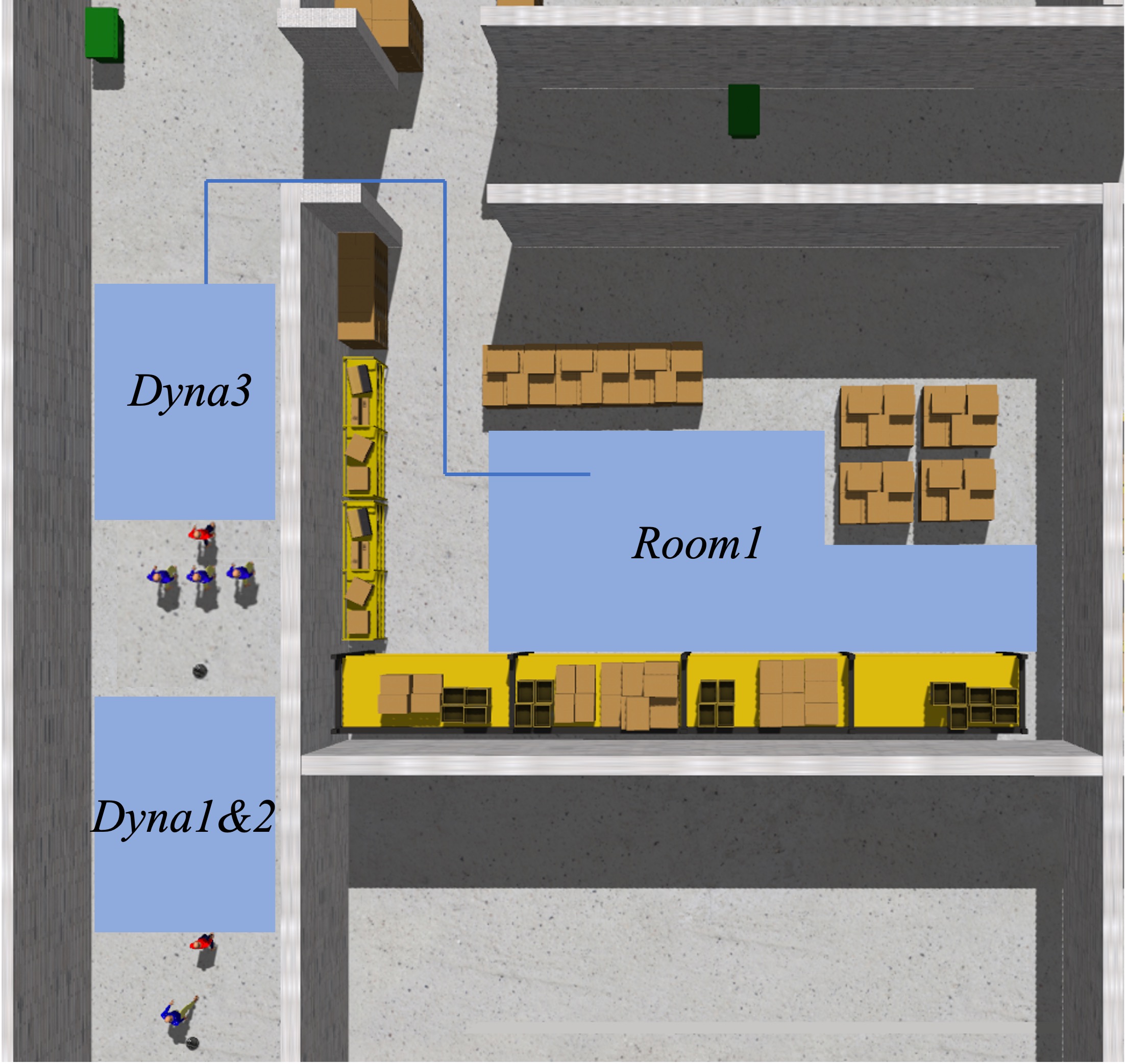}
        \caption{Dynamic occlusion}
        \label{fig:dynamicInFactory}
    \end{subfigure}%
    \\
    \begin{subfigure}[t]{0.28\textwidth}
        \centering
        \includegraphics[width=\textwidth]{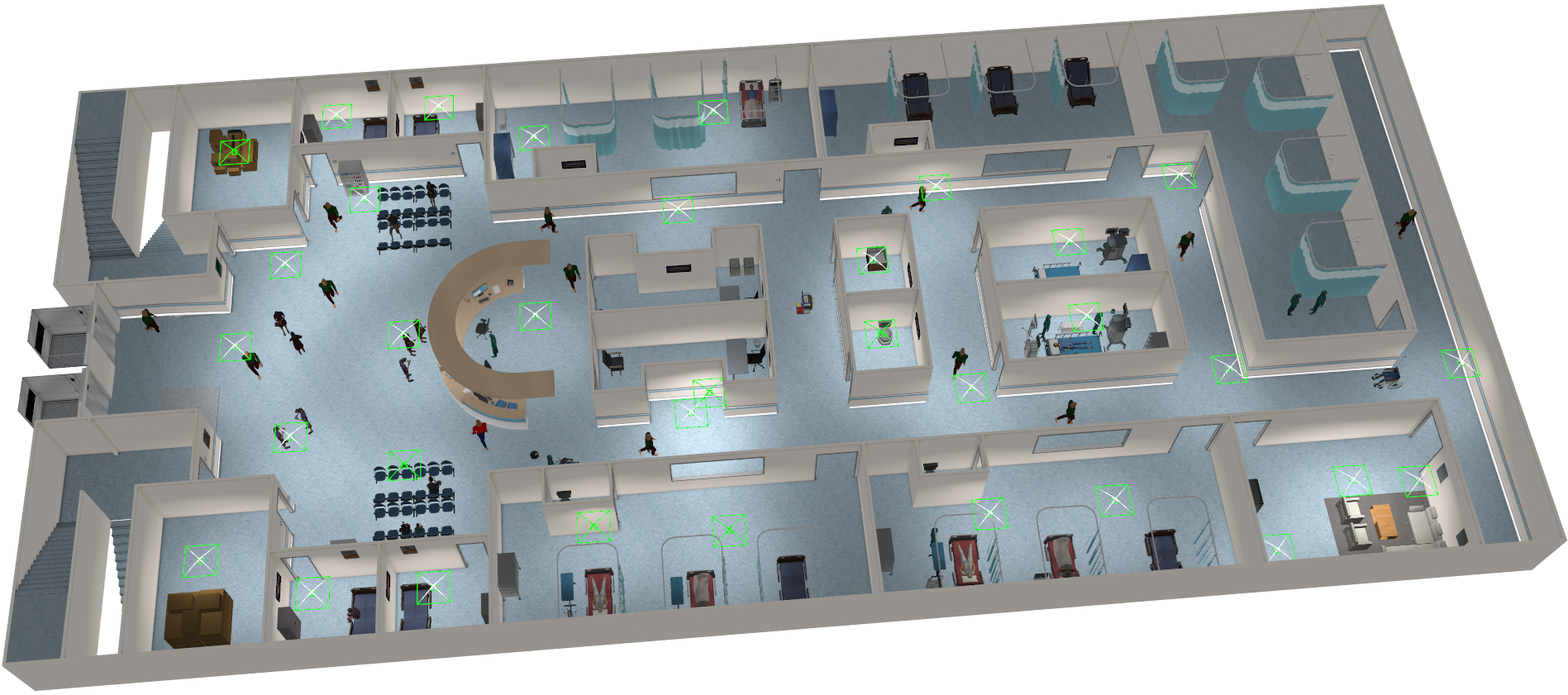}
        \caption{Hospital Env.}
        \label{fig:hospital}
    \end{subfigure}%
    \begin{subfigure}[t]{0.19\textwidth}
        \centering
        \includegraphics[width=\textwidth]{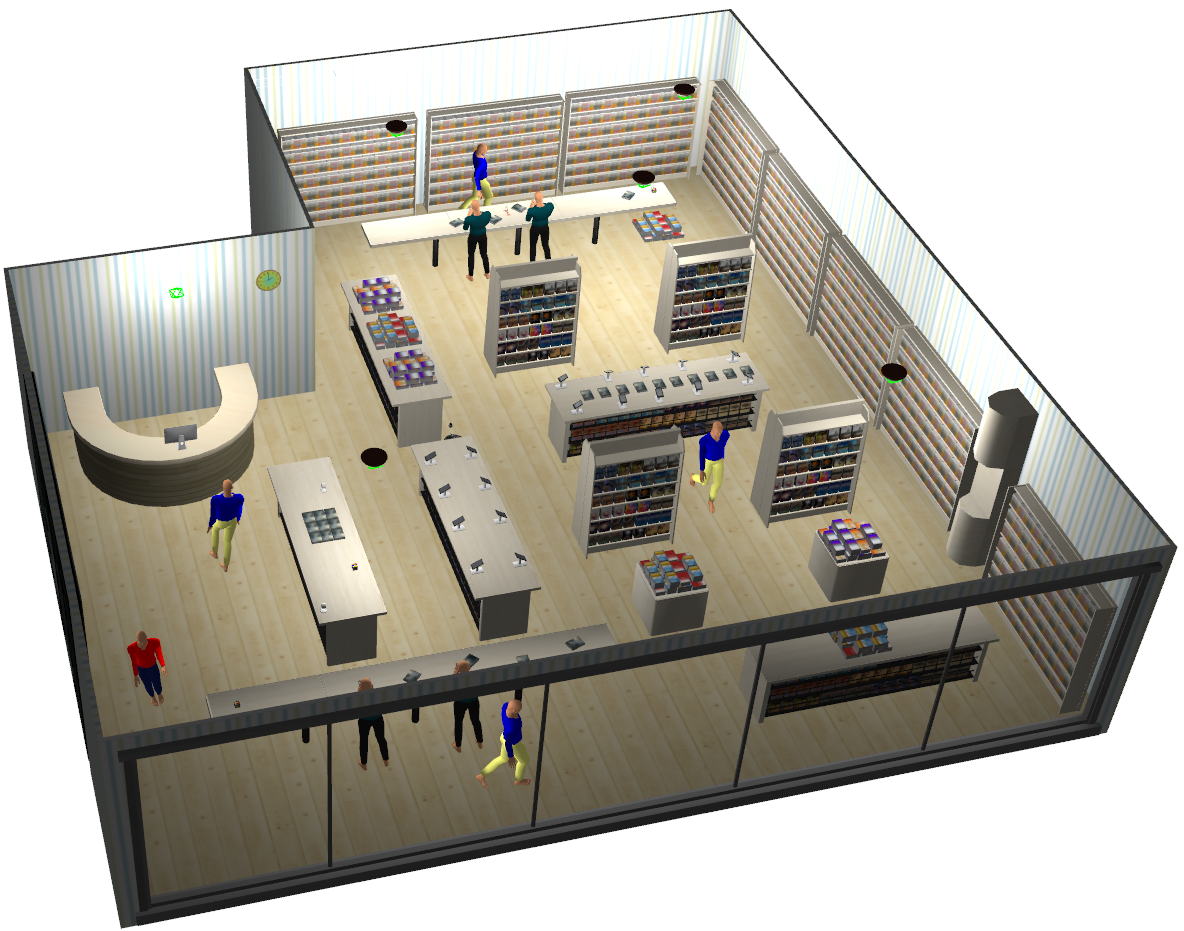}
        \caption{Bookstore Env.}
        \label{fig:bookstore}
    \end{subfigure}%
\caption{Simulated environments comprise three distinct scales and room structures: (a) \textbf{Factory} (large scale), (c) \textbf{Hospital} (medium scale), and (d) \textbf{Bookstore} (small scale). Six scenarios based on Factory are designed to independently assess the robot's person-search ability under topographic occlusion (\textit{Room1}, \textit{Room2} and \textit{Room3}) and dynamic occlusion (\textit{Dyna1}, \textit{Dyna2} and \textit{Dyna3}), as shown in (a) and (b), respectively. We conducted three dynamic occlusion experiments: \textit{Dyna1} involves a single occluder moving, \textit{Dyna2} features a single occluder with a gradually decreasing speed, and \textit{Dyna3} simulates scenarios with multiple occluders.}
\label{fig:simEnv}
\vspace{-15pt}
\end{figure}

We evaluated our method in both real-world and simulated environments to ensure its robustness. The simulated settings included three room layouts with different scales: \textbf{Bookstore} (small), \textbf{Hospital} (medium), and \textbf{Factory} (large), as shown in Figs. \ref{fig:bookstore}, \ref{fig:hospital}, and \ref{fig:factory}. These environments featured various objects and both static and dynamic individuals to create frequent occlusions, which are challenging for RPF. Additionally, within the factory, we designed six scenarios to assess the robot’s person-search capabilities under topographic and dynamic occlusions (see Figs. \ref{fig:factory} and \ref{fig:dynamicInFactory}). These scenarios include difficult U-turns and dynamic occlusions inspired by typical interactions, simulating complex real-life environments. Specifically, we developed three dynamic occlusion experiments: \textit{Dyna1}: simulates the single-person occlusion with a constant low speed (1.0~m/s). \textit{Dyna2}: features a single-person occlusion with decreasing speed (from 1.5~m/s to 0.8~m/s). \textit{Dyna3}: represents multiple-person occlusions. Due to the page limitation, the performance evaluations across different environment scales and room configurations have been placed in the Appendix.

In the simulated environments, we use a differential-drive robot with a 2D LiDAR for navigation and an RGB-D camera for perception. A target person is re-identified after being observed in over five consecutive frames. Locations are estimated using YOLOX object detection and depth data. Individuals interact according to the social force model \cite{helbing1995social}, ensuring realistic movements. The target's start and end positions are randomly generated within free space to create various scenarios. Real-world scenarios are illustrated in Fig.~\ref{fig:cover}. Due to page limitations, the supplementary materials present all real-world qualitative and quantitative results, the evaluations across different scales and room structures, and the complete implementation details.

\subsection{Compared Algorithms and Evaluation Metrics}
To validate the effectiveness of our proposed method, we compared it against several existing person-search baselines. The first baseline, \textbf{GoToLostLocation (GTLL)} \cite{chen2017integrating}, directs the robot to the target's last known location when lost and then rotates to search. The second baseline, \textbf{GoToPredictedLocation (GTPL)} \cite{Kim2018AnAF}, uses an SVR-based model to predict and navigate to the target's location before searching. 
Given the scarcity of methods targeting person search in unknown environments, we enhanced these baselines by incorporating a frontier map and a greedy next-best-view (NBV) strategy, resulting in \textbf{GTLL + Greedy-NBV} and \textbf{GTPL + Greedy-NBV}, where the robot first navigates to the initial location and then to the nearest frontier. We also integrated the map-based \textbf{HB-Particle} method \cite{goldhoorn2017search}, which directs the robot to the highest-belief particle with particles propagating into unknown regions, and the information gain-based \textbf{Active Graph-SLAM} \cite{placed2023ral}, which employs frontier-based exploration to reduce pose graph uncertainty.

To assess all methods, we used established person-search metrics: \textbf{Success Rate (SR)} \cite{goldhoorn2017search, Bayoumi2019search}, \textbf{Success Weighted by Inverse Path Length (SPL)} \cite{anderson2018evaluation}, and \textbf{Trajectory and Velocity Analysis (TVA)}.  SR measures the percentage of successful searches within a 180-second time limit over 100 runs per scenario. SPL evaluates search efficiency by considering both success and path optimality. TVA examines the smoothness of robot motion by plotting trajectories, velocities, and distances to the target in various dynamic occlusion scenarios.

\begin{figure*}[t]
    \centering
    \includegraphics[width=0.88\textwidth, trim={10pt 7pt 10pt 10pt}, clip]{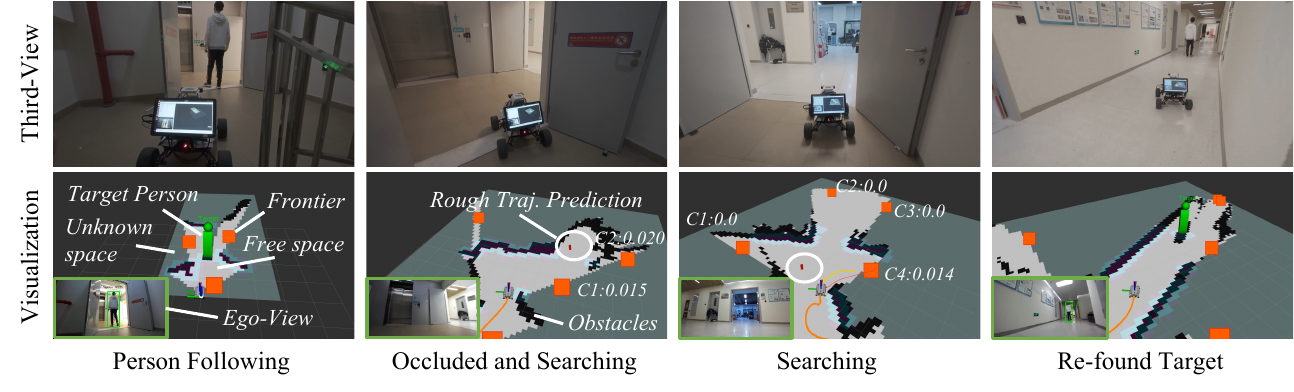}
    \caption{Searching keyframes of our method in topographic occlusion scenarios in real-world experiments. The first row shows the third view. The second row shows the explored occupancy map, frontiers (red squares), robot and target positions, ego view, and robot trajectories (orange lines). Once the occlusion happened, trajectory predictions direct the robot toward the region between the left corridor and the room, but our belief-guided graph updating integrates new environmental data and the target’s motion history to penalize unlikely left-corridor and room candidates to zero belief (candidate 1, 2 and 3) and focus the search on more promising areas towards the right corridor (candidate 4 with 0.014 belief) and finally re-found the target.}
    \label{fig:topoVisTopoOcc}
\vspace{-5pt}
\end{figure*}

\subsection{Ablation Experiments}

\subsubsection{Verification of the inference factor and probabilistic inheritance}
To assess the roles of the inference factor and probabilistic inheritance in our belief-guided search framework, we conduct an ablation study with results shown in Table~\ref{tab:ablationTopographic}. Removing either component can cause the robot to navigate incorrectly and significantly reduce its performance. For example, in \textit{Room3}, SR and SPL were 62\% and 60.8\%, and 71\% and 68.5\%, respectively. Similar declines were observed in \textit{Factory} and \textit{Hospital}. Visualizations of baselines and our method are shown in the supplementary materials.

Here, we present a visualization of our real-world experiments (see Fig.~\ref{fig:topoVisTopoOcc}). Initially, the trajectory prediction guided the robot toward the room and left corridor. However, the inference factor incorporated new environmental data, eliminating the left corridor and room candidates (1, 2, and 3) by reducing their belief to zero. This redirected the robot toward the higher-belief candidates in the right corridor (candidate 4 with a 0.014 belief), ensuring a successful search. Without the inference factor, the robot would have followed the initial trajectory, resulting in a failed search.

Furthermore, probabilistic inheritance enables new candidates to inherit beliefs from their parent candidates with decay. This allows the robot to focus on more probable candidates as the search progresses. As shown in the third and fourth subfigures of Fig.~\ref{fig:topoVisTopoOcc}, candidates in the right corridor continue to inherit belief, guiding the robot to successfully re-find the target. Without probabilistic inheritance, distant candidates would lose all belief, leading to incorrect navigation and search failure. Our approach leverages the local connectivity of frontiers, optimizing search direction towards likely locations. In conclusion, both the inference factor and probabilistic inheritance are essential for effective navigation and search in complex environments, as demonstrated by improved performance in our ablation study.

\begin{table}[t]
    \centering
    \caption{Ablation study for our belief-guided field in topographic occlusion scenarios. \textbf{``W/o inference factor''} represents our RPF-Search without the inference factor (Eq.~\ref{eq:inferenceFactor}) and \textbf{``w/o probabilistic inheritance''} excludes the probabilistic inheritance (Eq.~\ref{eq:probabilityInheritance}). Evaluation metrics include success rate (SR, \%) and success weighted by inverse path length (SPL, \%). We run 100 times for each scenario.}
    \scalebox{0.73}{ 
        \begin{tabular}{l|cc|cc|cc}
            \toprule 
            \multirow{2}{*}{\textbf{Method}} &\multicolumn{2}{c|}{\textbf{Factory$^*$(\%)}} &\multicolumn{2}{c|}{\textbf{Hospital$^*$(\%)}} &\multicolumn{2}{c}{\textbf{Room3}(\%)} \\
            & \textit{SR}\textuparrow & \textit{SPL}\textuparrow & \textit{SR}\textuparrow & \textit{SPL}\textuparrow & \textit{SR}\textuparrow & \textit{SPL}\textuparrow \\
            \midrule
            \midrule
            \textbf{w/o inference factor} &68 &62.0 &75 &67.0 &62 &60.8 \\
            \textbf{w/o probabilistic inheritance} &56 &46.4 &63 &54.3 &71 &68.5 \\
            
            \rowcolor{gray!25}
            \textbf{RPF-Search (ours)} &\bf 100 &\bf 94.7 &\bf 98 &\bf 94.8 &\bf 100 &\bf 96.2 \\
            \bottomrule
        \end{tabular}
    }
    \label{tab:ablationTopographic}
\vspace{-5pt}
\end{table}

\begin{table}[t]
    \centering
    \caption{Ablation study for our overtaking and fluid-following fields in dynamic occlusion scenarios. \textbf{W/o obs.-based overtaking field} refers to RPF-Search without the observation-field-based overtaking mechanism, while \textbf{w/o fluid-following field} excludes the fluid-following component.
    }
    \scalebox{0.70}{ 
        \begin{tabular}{l|cc|cc|cc}
            \toprule 
            \multirow{2}{*}{\textbf{Method}} &\multicolumn{2}{c|}{\textbf{Dyna1}(\%)} &\multicolumn{2}{c|}{\textbf{Dyna2}(\%)} &\multicolumn{2}{c}{\textbf{Dyna3}(\%)} \\
            & \textit{SR}\textuparrow & \textit{SPL}\textuparrow & \textit{SR}\textuparrow & \textit{SPL}\textuparrow & \textit{SR}\textuparrow & \textit{SPL}\textuparrow \\
            \midrule
            \midrule
            \textbf{w/o obs.-based overtaking field} &25 &24.4 &24 &23.2 &91 &87.1 \\
            \textbf{w/o fluid-following field} &95 &91.6 &86 &82.3 &82 &78.5 \\
            
            \rowcolor{gray!25}
            \textbf{RPF-Search (ours)} &\bf 97 &\bf 92.8 &\bf 100 &\bf 95.6 &\bf 92 &\bf 88.0\\
            \bottomrule
        \end{tabular}
    }
    \label{tab:ablationDynamic}
\vspace{-15pt}
\end{table}

\begin{figure*}[t]
    \centering
        \begin{subfigure}[t]{0.48\textwidth}
            \centering
            \includegraphics[width=\textwidth]{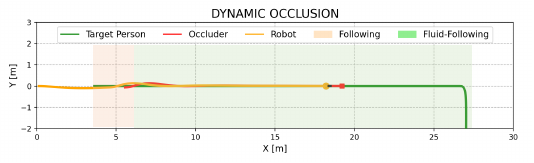}
            \caption{W/o overtaking (F, failed)}
            \label{fig:ablationVisDyna1a}
        \end{subfigure}%
        \begin{subfigure}[t]{0.48\textwidth}
            \centering
            \includegraphics[width=\textwidth]{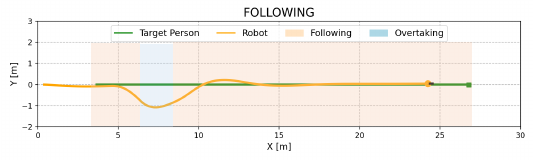}
            \caption{W/o fluid-following (S, succeed)}
            \label{fig:ablationVisDyna1b}
        \end{subfigure}%
        \vspace{0.004\linewidth}
        \begin{subfigure}[t]{0.48\textwidth}
            \centering
            \includegraphics[width=\textwidth]{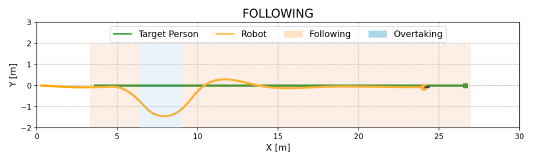}
            \caption{Ours (S, succeeded)}
            \label{fig:ablationVisDyna1c}
        \end{subfigure}%
        \begin{subfigure}[t]{0.48\textwidth}
            \centering
            \includegraphics[width=\textwidth]{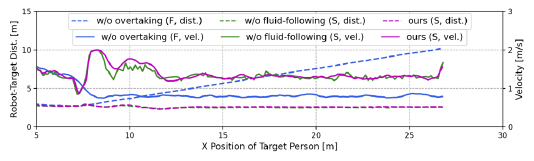}
            \caption{Comparison of robot-target distance and velocity across (a), (b), and (c)}
            \label{fig:ablationVisDyna1d}
        \end{subfigure}%
    \centering
    \caption{Visualization of the ablation study in \textit{Dyna1}, where a slow-moving distractor occludes the robot. \textbf{(a)} In the ``w/o overtaking'' scenario, the robot follows the occluder using fluid-following, increasing its distance from the target (blue dashed line in (d)), eventually failing to re-identify the target due to limited visibility. \textbf{(b)} In the ``w/o fluid-following'' case, the robot overtakes the occluder and quickly resumes tracking, maintaining a consistently close distance to the target (green dashed line in (d)). \textbf{(c)} In the ``ours'' scenario, the robot detects the slow-moving occluder. It performs an overtaking maneuver, leading to a successful search with a stable and close target distance (purple dashed line in (d)). \textbf{(d)} provides a comparison of (a), (b), and (c) in terms of robot-target distance (dashed lines) and velocity changes (solid lines).}
    \label{fig:ablationVisDyna1}
\vspace{-10pt}
\end{figure*}

\begin{figure*}[t]
    \centering
        \begin{subfigure}[t]{0.50\textwidth}
            \centering
            \includegraphics[width=\textwidth]{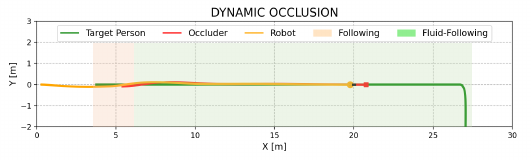}
            \caption{W/o overtaking (F, failed)}
            \label{fig:ablationVisDyna2a}
        \end{subfigure}%
        \hspace{0.05\linewidth}
        \begin{subfigure}[t]{0.40\textwidth}
            \centering
            \includegraphics[width=\textwidth]{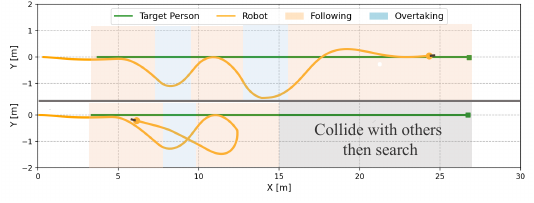}
            \caption{W/o fluid-following: top (S, succeed) and bottom (F, failed)}
            \label{fig:ablationVisDyna2b}
        \end{subfigure}%
        \vspace{0.004\linewidth}
        \begin{subfigure}[t]{0.48\textwidth}
            \centering
            \includegraphics[width=\textwidth]{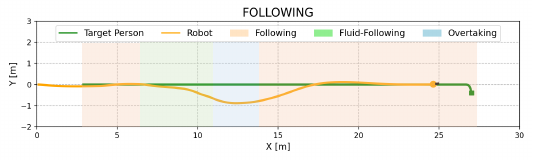}
            \caption{Ours (S, succeeded)}
            \label{fig:ablationVisDyna2c}
        \end{subfigure}%
        \begin{subfigure}[t]{0.48\textwidth}
            \centering
            \includegraphics[width=\textwidth]{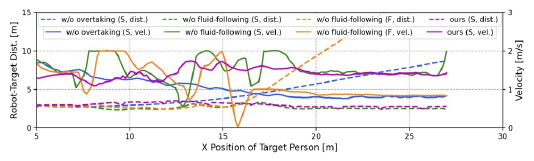}
            \caption{Comparison of robot-target distance and velocity}
            \label{fig:ablationVisDyna2d}
        \end{subfigure}%
    \centering
    \caption{Ablation study visualization in \textit{Dyna2} with decreasing occluder speed. In scenario \textbf{(a)} without overtaking, the robot trails the target, fails to re-identify them, and the robot-target distance increases (blue dashed line in (d)). In scenario \textbf{(b)} without fluid-following, two cases are shown: the top example where the robot eventually re-identifies the target but exhibits unstable motion due to frequent speed changes while overtaking a fast occluder (green solid line in (d)), and the bottom example where aggressive navigation leads to collisions and an increased distance (orange dashed line in (d)), resulting in search failure. In scenario \textbf{(c)}, our method adapts fluid-following when the occluder is fast and switches to overtaking as it slows, ensuring a smooth trajectory and stable motion (purple solid line in (d)). Panel \textbf{(d)} describes robot-target distance (dashed lines) and velocity changes (solid lines).}
    \label{fig:ablationVisDyna2}
\vspace{-10pt}
\end{figure*}

\begin{figure*}[t]
    \centering
        \begin{subfigure}[t]{0.50\textwidth}
            \centering
            \includegraphics[width=\textwidth]{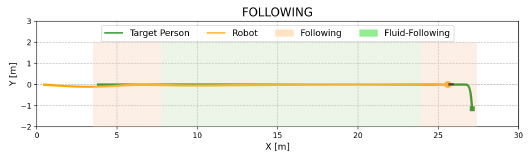}
            \caption{W/o overtaking (S, succeeded)}
            \label{fig:ablationVisDyna3a}
        \end{subfigure}%
        \hspace{0.05\linewidth}
        \begin{subfigure}[t]{0.40\textwidth}
            \centering
            \includegraphics[width=\textwidth, trim={0pt 0pt 0pt 1pt}, clip]{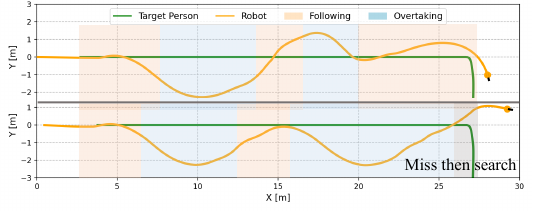}
            \caption{W/o fluid-following: top (S, succeed) and bottom (F, failed)}
            \label{fig:ablationVisDyna3b}
        \end{subfigure}%
        \vspace{0.004\linewidth}
        \begin{subfigure}[t]{0.48\textwidth}
            \centering
            \includegraphics[width=\textwidth]{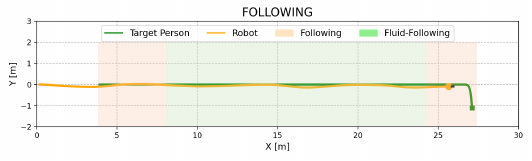}
            \caption{Ours (S, succeeded)}
            \label{fig:ablationVisDyna3c}
        \end{subfigure}%
        \begin{subfigure}[t]{0.48\textwidth}
            \centering
            \includegraphics[width=\textwidth]{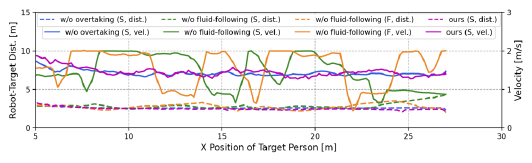}
            \caption{Comparison of robot-target distance and velocity}
            \label{fig:ablationVisDyna3d}
        \end{subfigure}%
    \centering
    \caption{Ablation study visualization in \textit{Dyna3} with multiple occluders. In scenario \textbf{(a)} without overtaking, the robot maintains stable fluid-following behavior, successfully resuming person-following when the target reappears, resulting in a close robot-target distance (blue dashed line) and steady motion (blue solid line) in (d). In scenario \textbf{(b)} without fluid-following, the robot attempts to overtake multiple occluders, leading to unstable motion and frequent shifts in the field of view, sometimes failing to relocate the target when the view diverges (bottom case). In scenario \textbf{(c)}, our method detects the absence of clear overtaking opportunities and continues fluid-following until the target becomes visible again, ensuring smooth motion (purple solid line) and maintaining a close robot-target distance (purple dashed line) in (d). Panel \textbf{(d)} describes robot-target distance (dashed lines) and velocity changes (solid lines)}
    \label{fig:ablationVisDyna3}
\vspace{-10pt}
\end{figure*}

\begin{table}[t]
    \centering
    \caption{Person-search performance was evaluated in our self-built three scenarios to independently assess search capabilities under topographic occlusion (\textbf{Room1}, \textbf{Room2}, and \textbf{Room3}). Each scenario was run 100 times. The best result is indicated in bold, while the second-best result is underscored.}
    \scalebox{0.76}{ 
        \begin{tabular}{l|cc|cc|cc}
            \toprule 
            \multirow{2}{*}{\textbf{Method}} & \multicolumn{2}{c|}{\textbf{Room1}} & \multicolumn{2}{c|}{\textbf{Room2}} & \multicolumn{2}{c}{\textbf{Room3}} \\
            & \textit{SR}\textuparrow & \textit{SPL}\textuparrow & \textit{SR}\textuparrow & \textit{SPL}\textuparrow & \textit{SR}\textuparrow & \textit{SPL}\textuparrow \\
            \midrule
            \midrule
            \textbf{GTLL\cite{chen2017integrating}} & 8 & 8.0 & 94 & 79.6 & 1 & 1.0 \\
            \textbf{GTPL\cite{Kim2018AnAF}} & 78 & 73.4 & 62 & 53.3 & 45 & 41.8 \\
            \textbf{GTLL\cite{chen2017integrating} + Greedy-NBV} & 77 & 74.1 & \underline{94} & \underline{80.5} & \underline{65} & 57.9 \\
            \textbf{GTPL\cite{Kim2018AnAF} + Greedy-NBV} & \underline{94} & \underline{84.6} & \underline{79} & \underline{68.0} & 42 & 40.2 \\
            \textbf{Active Graph-SLAM\cite{placed2023ral}} & 33 & 33.0 & 14 & 21.6 & 61 & \underline{60.8} \\
            \textbf{HB-Particle\cite{goldhoorn2017search}} & 2 & 2.0 & 14 & 12.2 & 0 & 0.0 \\
            \rowcolor{gray!25}
            \textbf{RPF-Search (ours)} & \bf 100 & \bf 96.1 & \bf 100 & \bf 97.1 & \bf 100 & \bf 96.2 \\
            \bottomrule 
        \end{tabular}
    }
    \label{tab:RoomResults}
\vspace{-10pt}
\end{table}

\subsubsection{Verification of two field-based strategies} 
We evaluated the effectiveness of the overtaking and fluid-following fields through an ablation study, with outcomes shown in Table~\ref{tab:ablationDynamic}. In \textit{Dyna1}, removing the fluid-following field slightly decreased the SR by 2\%, whereas removing the observation-based overtaking field caused a drop of 70\% in SR. Without overtaking, the robot followed a slow-moving occluder, increasing the target distance and failing to re-identify the target when it turned a corner (Fig.~\ref{fig:ablationVisDyna1a}). In contrast, the overtaking field enabled the robot to overtake the occluder, maintaining a stable and close distance to the target and successfully completing the search (Fig.~\ref{fig:ablationVisDyna1b}).

In \textit{Dyna2}, where the occluder's speed decreases from 1.5 m/s to 0.8 m/s, the absence of either the observation-based overtaking field or the fluid-following field resulted in reduced performance. Specifically, SR and SPL decreased by 76\% and 14\%, respectively, without the overtaking field, and by 72.4\% and 13.3\%, respectively, without the fluid-following field (Table~\ref{tab:ablationDynamic}). Simply following the occluder in this scenario increases the robot-target distance, risking the loss of the target (see Fig.~\ref{fig:ablationVisDyna2a}). Conversely, direct overtaking can lead to unstable motion or collisions, especially when pursuing a fast-moving occluder (see Fig.~\ref{fig:ablationVisDyna2b}). Our complete method addresses these issues by overtaking only when clear opportunities arise, such as when the overtaking cost is low and the occluder is moving slowly. This approach ensures smooth trajectories and stable motion, as demonstrated in Figs.~\ref{fig:ablationVisDyna2c} and ~\ref{fig:ablationVisDyna2d}.

In \textit{Dyna3}, involving multiple occluders, the absence of the fluid-following field led to frequent and unstable overtaking attempts, resulting in unstable motion and frequent velocity changes (see Fig.~\ref{fig:ablationVisDyna3b}). In contrast, our complete method employs a fluid-following strategy when overtaking is not feasible, ensuring smooth navigation and successful recovery once the target reappears after multi-person occlusion (refer to Figs.~\ref{fig:ablationVisDyna3c} and~\ref{fig:ablationVisDyna3d}). Overall, the overtaking field is essential for maintaining target tracking, while the fluid-following field ensures safe and stable navigation when overtaking is not possible. The combination of both fields significantly enhances person-search ability under dynamic occlusions.

\begin{table}[t]
    \centering
    \caption{Person-search performance in dynamic occlusion scenarios.}
    \scalebox{0.76}{ 
        \begin{tabular}{l|cc|cc|cc}
            \toprule 
            \multirow{2}{*}{\textbf{Method}} & \multicolumn{2}{c|}{\textbf{Dyna1}} & \multicolumn{2}{c|}{\textbf{Dyna2}} & \multicolumn{2}{c}{\textbf{Dyna3}} \\
            & \textit{SR}\textuparrow & \textit{SPL}\textuparrow & \textit{SR}\textuparrow & \textit{SPL}\textuparrow & \textit{SR}\textuparrow & \textit{SPL}\textuparrow \\
            \midrule
            \midrule
            \textbf{GTLL\cite{chen2017integrating}} & 0 & 0.0 & 0 & 0.0 & 0 & 0.0 \\
            \textbf{GTPL\cite{Kim2018AnAF}} & 8 & 8.0 & 5 & 4.9 & 0 & 0.0 \\
            \textbf{GTLL\cite{chen2017integrating} + Greedy-NBV} & \underline{51} & \underline{48.8} & \underline{79} & \underline{75.7} & 10 & 9.9 \\
            \textbf{GTPL\cite{Kim2018AnAF} + Greedy-NBV} & 42 & 41.3 & 29 & 28.8 & \underline{34} & \underline{33.3} \\
            \textbf{Active Graph-SLAM\cite{placed2023ral}} & 4 & 4.0 & 1 & 1.0 & 1 & 1.0 \\
            \textbf{HB-Particle\cite{goldhoorn2017search}} & 34 & 33.2 & 12 & 11.5 & 1 & 0.9 \\
            \rowcolor{gray!25}
            \textbf{RPF-Search (ours)} & \bf 97 & \bf 92.8 & \bf 100 & \bf 95.6 & \bf 92 & \bf 88.0 \\
            \bottomrule 
        \end{tabular}
    }
    \label{tab:DynaResults}
\vspace{-10pt}
\end{table}

\subsection{Topological occlusion scenarios}
To independently evaluate the person-search capabilities of both the baseline methods and our approach, we designed three factory scenarios (\textit{Room1}, \textit{Room2}, and \textit{Room3}) focusing on topographic occlusion challenges, including regular paths and U-turns. As shown in Table~\ref{tab:RoomResults}, our method achieved the highest SR and SPL in all rooms: 100\% SR and 96.1\% SPL in \textit{Room1}, 100\% SR and 97.1\% SPL in \textit{Room2}, and 100\% SR and 96.2\% SPL in \textit{Room3}. These results outperform the second-best methods by up to +35\% in SR and +35.4\% in SPL, demonstrating that our approach is more effective in re-identifying targets and enables robots to follow near-optimal paths, reducing unnecessary movements and search time.

Among the baselines, ``GTPL + Greedy-NBV'' performed well in \textit{Room1} with 94\% SR and 84.6\% SPL by accurately predicting target locations in simpler environments. ``GTLL + Greedy-NBV'' achieved 94\% SR and 80.5\% SPL in \textit{Room2} since the lost target is usually within the room, allowing the greedy search to effectively locate the person. However, in \textit{Room3}, limited historical information led ``GTPL + Greedy-NBV'' to predict trajectories toward the corridor, misguiding the robot towards irrelevant exploration frontiers; consequently, it results in a greedy search away from the target room and achieves only 42\% SR and 40.2\% SPL. In addition, ``GTLL + Greedy-NBV'' reached low performance with 65\% SR and 57.9\% SPL in \textit{Room1} since it solely relies on the target's last known position without effective predictive capabilities.

In contrast, our method transforms predicted trajectories into probabilistic representations and uses a belief-guided search field that dynamically updates the estimated location by incorporating new environmental data and the target's past motion cues. This approach allows efficient searching and locating of the person, achieving 100\% SR and 96.2\% SPL in \textit{Room3}, significantly outperforming the baseline methods.

\subsection{Dynamic Occlusion Scenarios}
Dynamic occlusions are common in everyday person-following tasks, especially in environments crowded with people. While robots can often recover tracking after short-term occlusions—since the target person reappears quickly—the greatest challenge arises with long-term occlusions, where other pedestrians obstruct the target for extended periods. As introduced in Fig.~\ref{fig:dynamicInFactory}, we designed three dynamic occlusion experiments for simulating realistic long-term occlusion scenarios including \textit{Dyna1}, \textit{Dyna2} and \textit{Dyna3}.
In these scenarios, the target person follows a fixed path into \textit{Room1} and is occluded for an extended period in the corridor by other individuals. The optimal search behavior should consider both search efficiency and safe navigation, enabling an intelligent switch between overtaking and following the occluders.

As shown in Table~\ref{tab:DynaResults}, our method achieves superior performance across all dynamic occlusion scenarios. Specifically, we obtained 97\% SR and 92.8\% SPL in \textit{Dyna1}, 100\% SR and 95.6\% SPL in \textit{Dyna2}, and 92\% SR and 88.0\% SPL in \textit{Dyna3}. Existing person-search methods do not explicitly address dynamic occlusion problems and, consequently, perform poorly in these environments. Our method surpasses the second-best baseline by: \textit{Dyna1}: +46\% SR and +44.0\% SPL, \textit{Dyna2}: +21\% SR and +19.9\% SPL, \textit{Dyna3}: +58\% SR and +54.7\% SPL.

\section{conclusion}
\label{sec:conclusion}

In this paper, we introduce RPF-Search, a novel heuristic-guided, field-based person-search framework designed for robust robot person-following in dynamic and unknown environments. This approach addresses key challenges in human-robot interaction, enabling robots to reliably locate a target person despite frequent occlusions. By constructing an environmental map in real-time and leveraging the motion cues of detected individuals, RPF-Search effectively adapts to both topographic occlusions—utilizing a belief-guided search field—and dynamic occlusions through an observation-based potential field combined with fluid-field dynamics. Experimental results in both simulated and real-world scenarios demonstrate RPF-Search's significant improvements in search efficiency and success rates compared to traditional person-search methods. This advancement offers a practical and reliable solution for personal assistance and human-robot collaboration. Future work could further enhance these results by incorporating advanced trajectory prediction models and exploring the use of semantic cues.





\bibliographystyle{IEEEtran}
\bibliography{ref}

\begin{IEEEbiography}[{\includegraphics[width=1.0in,height=1.25in,clip,keepaspectratio]{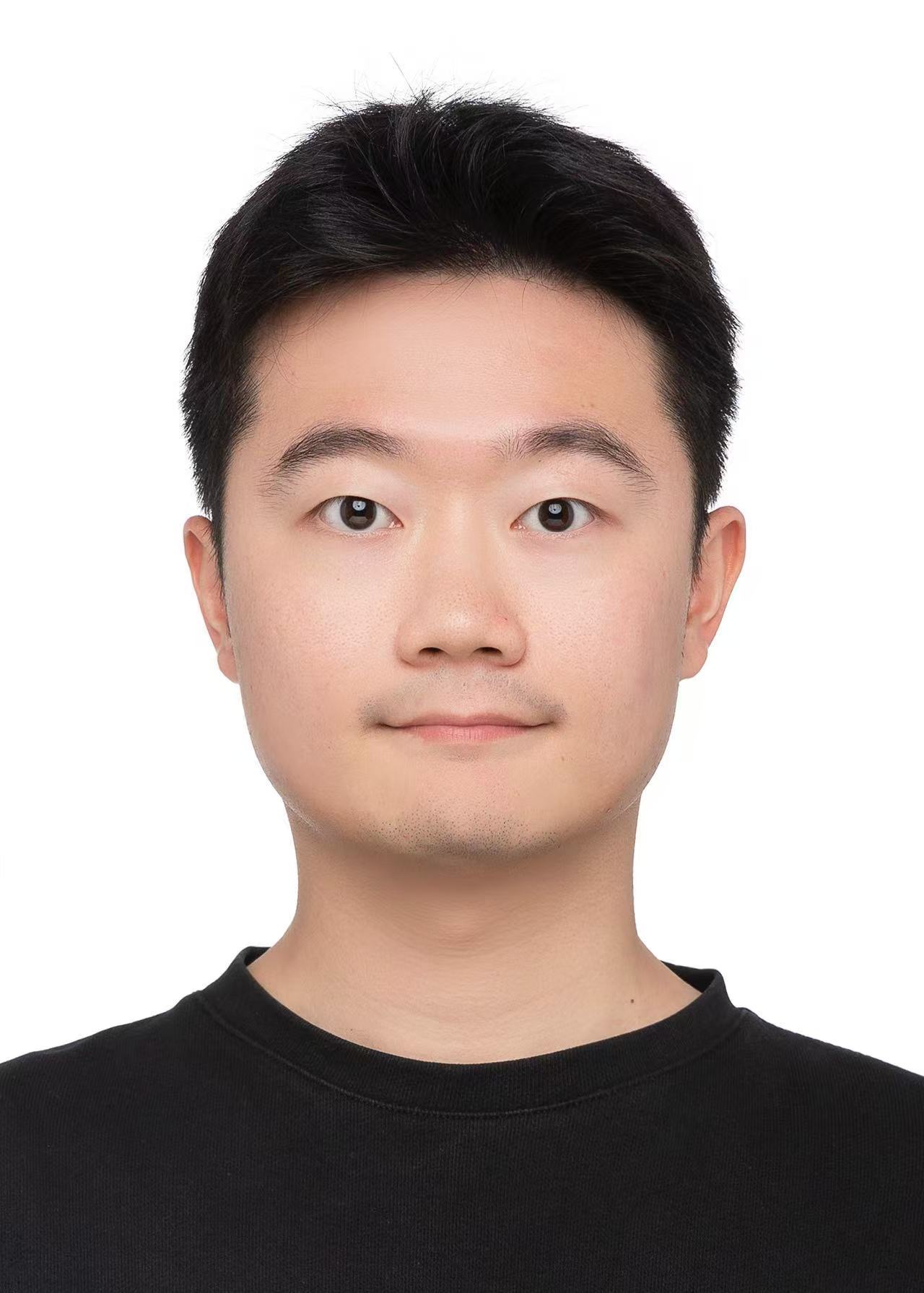}}]{Hanjing Ye}
received his B.E. and M.S. degrees from Guangdong University of Technology in 2019 and 2022, respectively. He is currently pursuing his Ph.D. degree in Robotics at the Southern University of Science and Technology. His research interests involve robot person following, autonomous navigation and human-robot interaction.\end{IEEEbiography}

\vspace{-45pt}

\begin{IEEEbiography}[{\includegraphics[width=1.0in,height=1.25in,clip,keepaspectratio]{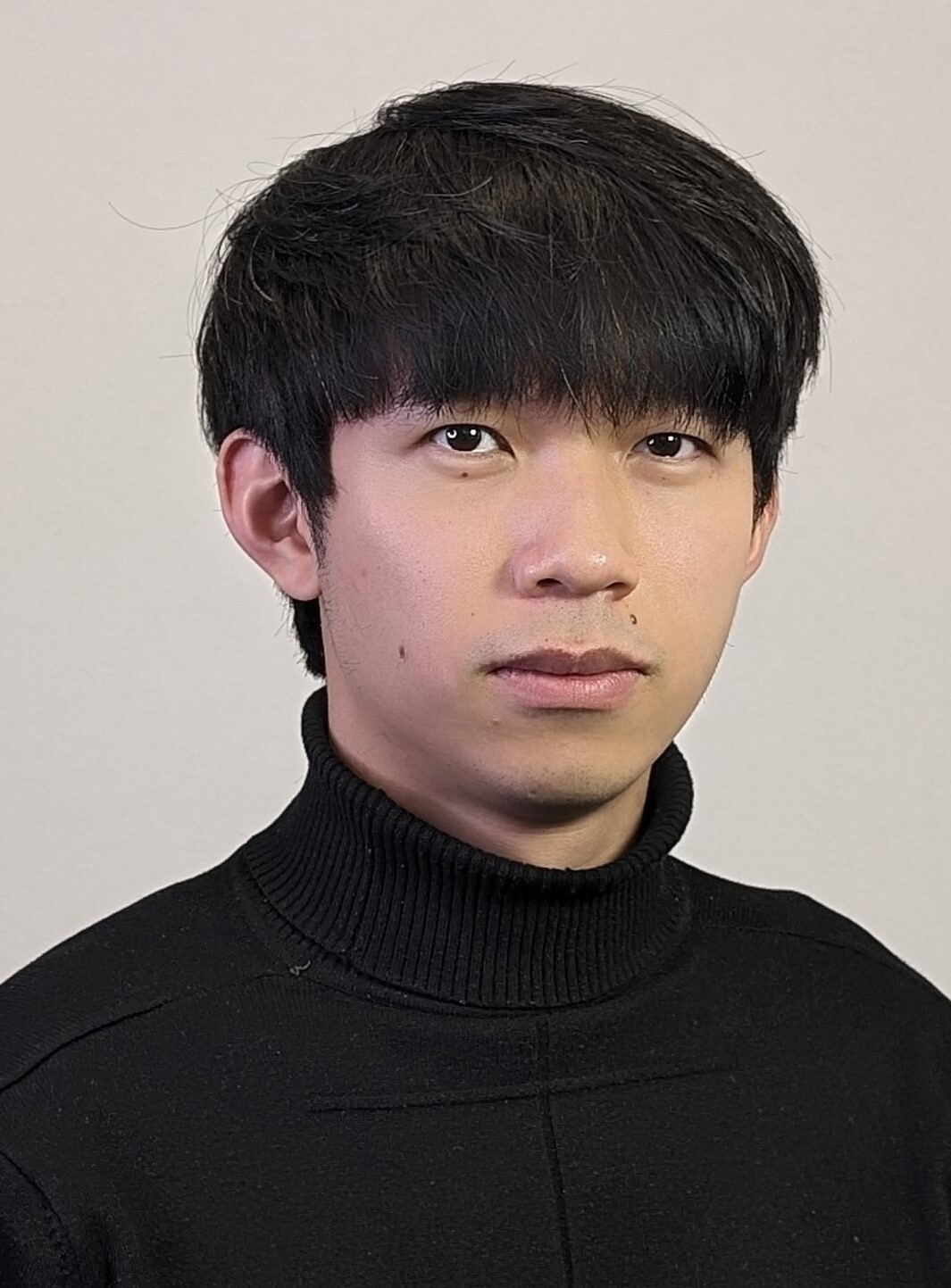}}]{Kuanqi Cai} received the B.E. degree in Transportation from Hainan University, Haikou, China, in 2018, and the M.E. degree in Mechanical Engineering from the Harbin Institute of Technology, Harbin, China, in 2021. He is currently pursuing the Ph.D. degree in Robotics at the Italian Institute of Technology and the Swiss Federal Institute of Technology in Lausanne (EPFL), Switzerland. From 2023 to 2024, he was a Research Associate at the Technical University of Munich, Germany. In 2021, he was honored as a Robotics Student Fellow at ETH Zurich. His research interests include motion planning and human–robot interaction.

\end{IEEEbiography}

\vspace{-45pt}

\begin{IEEEbiography}[{\includegraphics[width=1.0in,height=1.25in,clip,keepaspectratio]{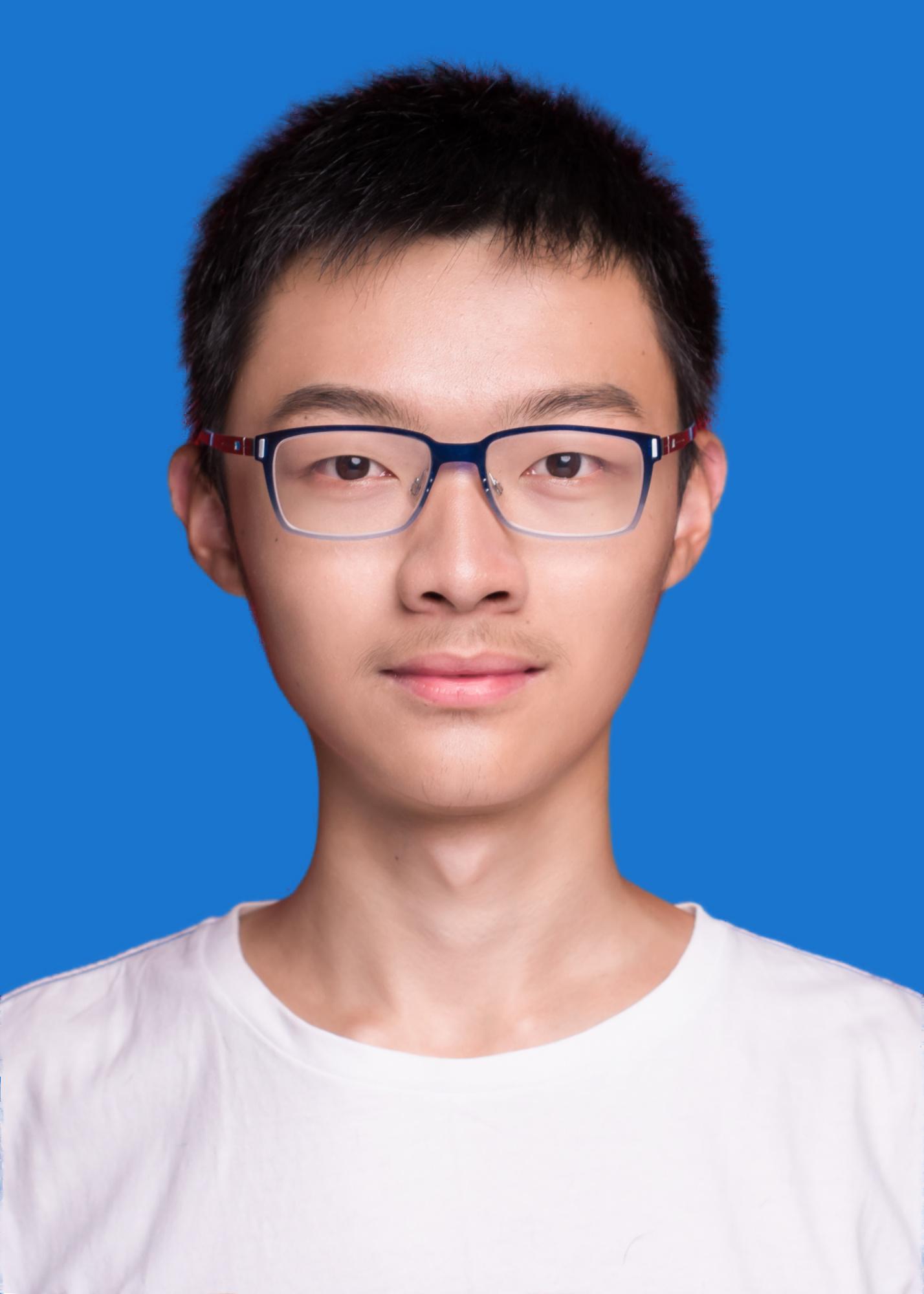}}]{Yu Zhan}
received his B.E. degree in microelectronics from Wuhan University in 2022. He is currently pursuing his Master’s degree from the Southern University of Science and Technology. His research interests include human pose estimation, human-robot interaction and robot vision.
\end{IEEEbiography}

\vspace{-45pt}

\begin{IEEEbiography}[{\includegraphics[width=1.0in,height=1.25in,clip,keepaspectratio]{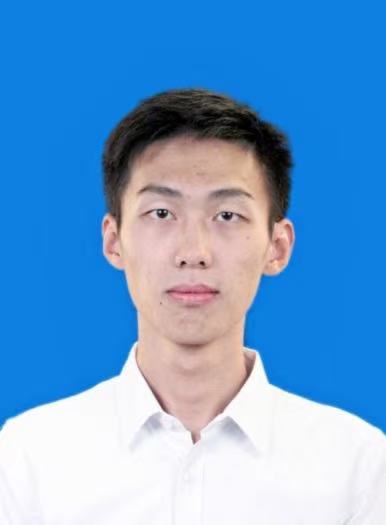}}]{Bingyi Xia}
received his B.E. and M.S. degrees from the Southern University of Science and Technology in 2020 and 2023, respectively. He is currently pursuing his Ph.D. degree with the Department of Electronic and Electrical Engineering at Southern University of Science and Technology, Shenzhen, China. His research interests include robot motion planning.
\end{IEEEbiography}

\vspace{-45pt}

\begin{IEEEbiography}[{\includegraphics[width=1.0in,height=1.25in,clip,keepaspectratio]{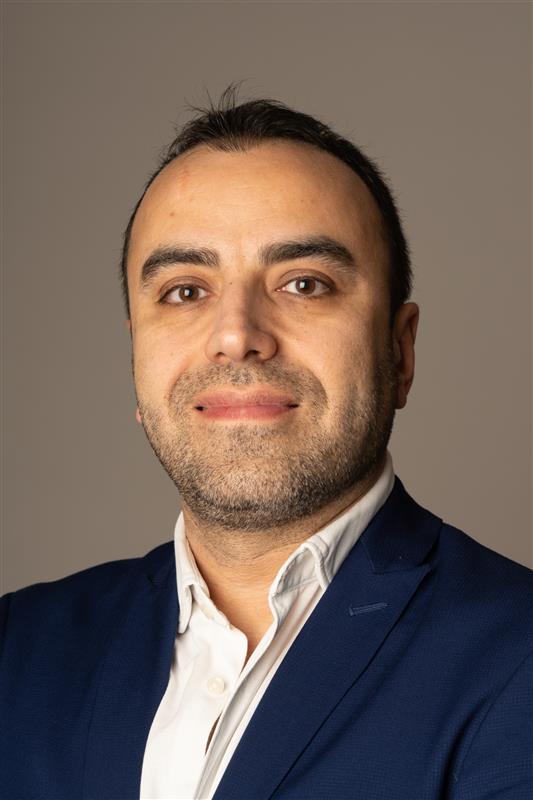}}]{Arash Ajoudani} received the M.D. degree in Biomedical Engineering from the Iran University of Science and Technology, Tehran, Iran, in 2007, and the Ph.D. degree in Robotics, Automation, and Bioengineering from the University of Pisa, Pisa, Italy, in 2014. He is the Director of the Human–Robot Interfaces and Interaction (HRI) Laboratory at the Istituto Italiano di Tecnologia, Genoa, Italy. His research interests include physical human–robot interaction, mobile manipulation, and adaptive control. Dr. Ajoudani was the recipient of the ERC grants Ergo-Lean in 2019 and Real-Move in 2023. He has received numerous awards, including the IEEE Robotics and Automation Society (RAS) Early Career Award in 2021 and the KUKA Innovation Award in 2018. He is a Senior Editor for The International Journal of Robotics Research (IJRR) and served as an elected member of the IEEE RAS Administrative Committee (AdCom) from 2022 to 2024. He has also coordinated multiple Horizon 2020 and Horizon Europe projects.
\end{IEEEbiography}

\vspace{-40pt}

\begin{IEEEbiography}[{\includegraphics[width=1.0in,height=1.25in,clip,keepaspectratio]{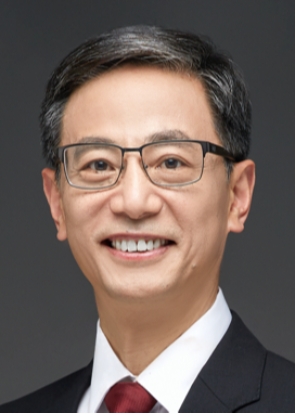}}]{Hong Zhang}
received his PhD in Electrical Engineering from Purdue University in 1986. He was a Professor in the Department of Computing Science, University of Alberta, Canada for over 30 years before he joined the Southern University of Science and Technology (SUSTech), China, in 2020, where he is currently a Chair Professor.  Dr. Zhang served as the Editor-in-Chief of the IROS Conference Paper Review Board (2020-2022) and is currently a member of the IEEE Robotics and Automation Society Administrative Committee (2023-25). He is a Fellow of IEEE and a Fellow of the Canadian Academy of Engineering.  His research interests include robotics, computer vision, and image processing.
\end{IEEEbiography}

\clearpage
\section*{APPENDIX}

\subsection{Methodology Supplements}
We additionally provide the method flows of our belief-guided search field and observation-based search field as shown in Algorithm.~\ref{alg:topoGraphicSearch} and Algorithm.~\ref{alg:dynamicOcclusionSearch}, respectively. In this paper, our frontier detection is based on \cite{placed2023ral}, which is designed to select the most informative frontiers for constructing a fully explored map. For Canny edge detection, we set the low and high thresholds to 0 and 255, respectively. Subsequently, we perform contour extraction to derive the boundaries of regions of interest. Finally, the centroids of these contours are determined and utilized as frontiers.
\begin{algorithm}[h]
    \DontPrintSemicolon
    \SetNoFillComment
    \footnotesize
    \KwIn{Occupancy map $\mathcal{M}_t$, Target person's historical trajectory $\mathcal{H}_t$, Robot position in the map frame $\mathbf{p}_{r,t}$}
    \KwOut{Optimal search goal $\mathbf{x}^{*}$}

    $\{\mathcal{F}_i\}_t=FrontierDetection(\mathcal{M}_t)$;\\
    \textcolor{blue}{\# For probabilistic inheritance}\\
    $Parent(\mathcal{F}_{i,t})=FindParent(\mathcal{F}_{i,t}), Parent(\mathcal{F}_{i,t}) \in \{\mathcal{F}_i\}_{t-1}$;\\
    
    \textcolor{blue}{\# For target person existence} \\
    $\{\bar{\mathbf{u}}_{i}\}^N_{i=1}=TrajectoryPrediction(\mathcal{H}_t)$;\\
    $\mathcal{V}_e(\cdot)=ProbPropagation(\{\bar{\mathbf{u}}_{i}\}^N_{i=1})$;\\

    \textcolor{blue}{\# Candidate weight calculation}\\
    \For {$\mathcal{F}_{i,t} \in \{\mathcal{F}_i\}_t$} {
    Inference factor $\eta_i = Simulation(\mathcal{H}_t[-1], \mathcal{F}_{i,t})$ $\longrightarrow$ Eq.~\ref{eq:inferenceFactor};\\
    Information gain $b_g = \mathcal{V}_g(\mathcal{F}_{i,t}, \mathcal{M}_t)$ $\longrightarrow$ Eq.~\ref{eq:informationGain};\\
    Collision risk $b_c = \mathcal{V}_c(\mathcal{F}_{i,t}, \mathcal{M}_t)$ $\longrightarrow$ Eq.~\ref{eq:collisionRisk};\\
    Target person existence $b_e = \mathcal{V}_e(\mathcal{F}_{i,t})$ $\longrightarrow$ Eq.~\ref{eq:targetPersonExistence};\\
    Probabilistic inheritance $b_p = \mathcal{V}_p(\mathcal{F}_{i,t}, Parent(\mathcal{F}_{i,t}))$ $\longrightarrow$ Eq.~\ref{eq:probabilityInheritance};\\
    $W_{i,t} = \eta_i(b_c+b_g+b_e+b_p)$ $\longrightarrow$ Eq.~\ref{eq:objectiveFunction};\\
    }
    ${\mathbf{x}^{*}} = \arg \max_{\mathbf{x}} \left\{ W_i \exp\left( -\frac{\| \mathbf{x} - \mathcal{F}_i \|^2}{2 \sigma^2} \right) \right\}$ $\longrightarrow$ Eq.~\ref{eq:beliefField};\\
    
    \textbf{Return} $\mathbf{x}^{*}$
\caption{Belief-guided search field}
\label{alg:topoGraphicSearch}
\end{algorithm}

\begin{algorithm}[h]
    \DontPrintSemicolon
    \SetNoFillComment
    \footnotesize
    \KwIn{Occupancy map $\mathcal{M}_t$, Occluders historical trajectory $\{\mathcal{H}_{{\rm occ},i}\}_t$, Robot position in the map frame $\mathbf{p}_{r,t}$, Robot velocity $v_{\rm robot}$, `Viscosity' of the fluid flow $\gamma$, Velocity thresholds $\delta_{v_{\text{min}}}$ and $\delta_{v_{\text{max}}}$, Cost threshold $\delta_c$}
    \KwOut{Optimal search goal $\mathbf{x}^*$}

    $\{\mathbf{x}_i\}=Sample(\mathcal{M}_t, \{\mathcal{H}_{{\rm occ},i}\}_t)$;\\
    Occluders' velocity $\{v_{{\rm occ},i}\}_t=CalVelocity(\{\mathcal{H}_{{\rm occ},i}\}_t)$;\\
    \textcolor{blue}{\# Observation-based potential field overtaking} \\
    $\{\bar{\mathbf{p}}_{i,j}\}^N_{j=1}=TrajectoryPrediction(\{\mathcal{H}_{{\rm occ},i}\}_t)$;\\
    $\mathbf{F}_{\text{rep}}(\cdot)=CalRepulsiveForce(\{\bar{\mathbf{p}}_{i,j}\}^N_{j=1},\mathcal{M}_t)$;\\
    $\mathbf{F}_{\text{att}}(\cdot)=CalAttractiveForce(\{\bar{\mathbf{p}}_{i,j}\}^N_{j=1}, \mathbf{p}_{r,t})$;\\
    $\left( \mathbf{x}^*, C^* \right) = \left( \mathop{\arg\min}_{\{\mathbf{x}_i\}}, \min \right) \left( \|\mathbf{F}_{\text{att}}(\mathbf{x}_i)\| + \|\mathbf{F}_{\text{rep}}(\mathbf{x}_i)\| \right)$;\\
    
    \uIf {$\max(\{v_{{\rm occ},i}\}_t) < \delta_{v_{\text{min}}}$ and $C^* > \delta_c$} {
        \textcolor{blue}{\# Switch to belief-guided searching} \\
        Get $\mathbf{x}^*$ from Algorithm 1;\\
    }
    \uElseIf {$\min(\{v_{{\rm occ},i}\}_t) > \delta_{v_{\text{max}}}$ or $C^* > \delta_c$} {
        \textcolor{blue}{\# Fluid field-based following} \\
        $\mathrm{v}(\cdot)=CalVelocityField(\{v_{{\rm occ},i}\}_t, \{\mathcal{H}_{{\rm occ},i}\}_t)$;\\
        $\rho_(\cdot)=CalDensityField(\{\mathcal{H}_{{\rm occ},i}\}_t)$;\\
        $\mathbf{x}^* = \mathop{\arg\min}_{\{\mathbf{x}_i\}} (\gamma \rho(\mathbf{x}_i) \cdot (\mathrm{v}_{\text{robot}} - \mathrm{v}(\mathbf{x}_i)))$;\\
    }
    \textbf{Return} $\mathbf{x}^*$
\caption{Observation-based search field}
\label{alg:dynamicOcclusionSearch}
\end{algorithm}

\subsection{Implementation Details and Metrics}
\begin{table}[t]
    \centering
    \caption{Parameters of simulated and real-world environments.}
    \scalebox{0.48}{
    \begin{tabular}{l|ccccccccc}
    \toprule 
    \multirow{2}{*}{\textbf{Simulated Environment}} &\multicolumn{7}{c}{\textbf{Factory}}  &\multirow{2}{*}{\textbf{Hospital}} &\multirow{2}{*}{\textbf{Bookstore}}\\
    &\textit{Room1} &\textit{Room2} &\textit{Room3} &\textit{Dyna1} &\textit{Dyna2} &\textit{Dyna3} &\textit{Factory$^*$} & & \\
    \midrule
    \midrule
    Size (m$\times$m) &\multicolumn{7}{c}{40$\times$70} &26$\times$50 &14$\times$13 \\
    Num. of people &1 &1 &1 &9 &9 &11 &20 &20 &9 \\
    Target's path length (m) &40.0 &55.0 &50.0 &57.0 &57.0 &57.0 &48.3 &28.6 &17.9\\
    Robot &\multicolumn{9}{c}{Kobuki differential robot with maximum 2.0 m/s}\\
    Sensors &\multicolumn{9}{c}{RGB-D camera (120$^{\degree}$ FoV, 0.03-10 m) and 2D laser rangefinder (0-20 m)}\\
    \midrule
    \midrule
    \textbf{Real-World Environment} &\multicolumn{3}{c}{\textbf{Large U-turn}} &\multicolumn{3}{c}{\textbf{Long Hallway (S)}} &\multicolumn{3}{c}{\textbf{Long Hallway (M)}}\\
    \midrule
    \midrule
    Size (m$\times$m) &\multicolumn{3}{c}{20$\times$3} &\multicolumn{3}{c}{50$\times$3} &\multicolumn{3}{c}{50$\times$3}\\
    Num. of people &\multicolumn{3}{c}{1} &\multicolumn{3}{c}{2} &\multicolumn{3}{c}{4}\\
    Target's path length (m) &\multicolumn{3}{c}{15} &\multicolumn{3}{c}{42} &\multicolumn{3}{c}{42}\\
    Robot &\multicolumn{9}{c}{Scout-Mini differential robot with maximum 1.8 m/s}\\
    Sensors &\multicolumn{9}{c}{Stereo camera (120$^{\degree}$ FoV, 0.03-10 m)}\\
    \bottomrule
    \end{tabular}
    }
    \label{tab:envParams}
\end{table}

\begin{table}[t]
    \centering
    \caption{Parameters of the proposed method.}
    \scalebox{0.7}{
    \begin{tabular}{l|cccccccc}
    \toprule 
    \textbf{Parameter} &$v_{\text{max}}$ &$\lambda$ &$d_s$ &$d_0$ &$\eta_{f}$ &$\eta_{a}$ &$\delta_{c}$ &$\delta_{v_{\text{min}}}$ \\
    \midrule
    \textbf{Value} &1.5 m/s &0.01 &1.5 m &1.3 m &1.1 &0.9 &10.0 &1.3 m/s\\
    \bottomrule
    \end{tabular}
    }
    \label{tab:hyperParams}
\vspace{-10pt}
\end{table}
The parameters of the proposed method are shown in Table.~\ref{tab:hyperParams}. In the simulated environments, our differential-drive robot is equipped with a 2D laser rangefinder (0-20 m) for navigation and an RGB-D camera (120° FoV, 0.03-10 m) for perception. The robot's maximum velocity is 2.0 m/s, and the simulated people's speeds range from 0.8 to 1.7 m/s. In our real-world experiments, we employ a Scout-Mini differential-drive robot equipped with a stereo camera that offers a 120° horizontal field of view (FoV) and a range of 0.03–10 meters. The stereo camera provides both RGB and depth images. To maintain consistency with our simulations, all planners and controllers are configured identically. For target person tracking and re-identification, we adopt the vision-based tracking method proposed in \cite{ye2023reid}. All parameters involving simulated and real-world environments are shown in Table~\ref{tab:envParams}.

We additionally evaluate the following metric: \textbf{Euclidean Distance to Target Over Time (EDTOT)}: An effective person-search approach should help the robot quickly recover the person-following, resulting in a smaller average distance to the target person, so we measure the Euclidean distance between the robot and the target person in each episode.

\subsection{Real-world Experiments} 
\begin{figure}
    \centering
    \includegraphics[width=0.85\linewidth, trim={0pt 80pt 0pt 0pt}, clip]{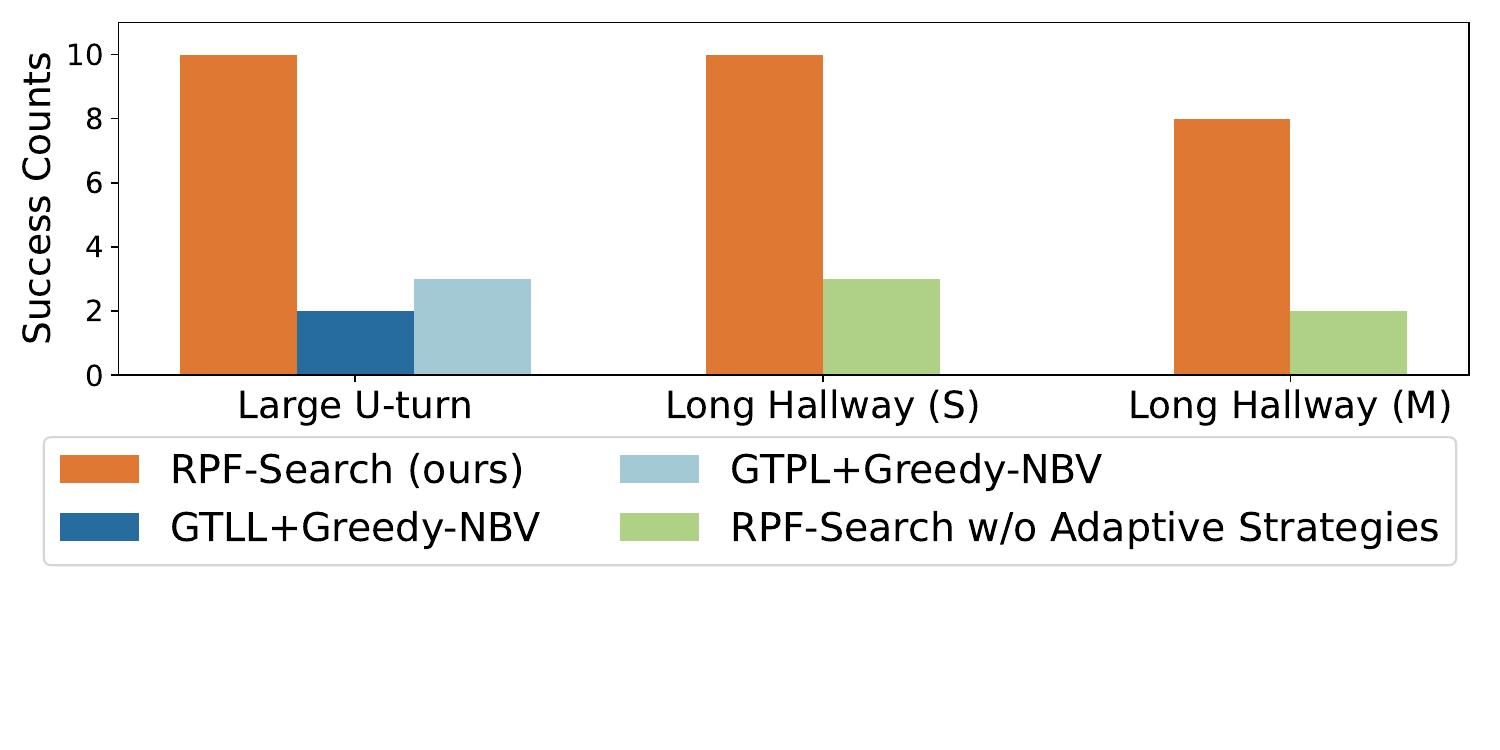}
    \caption{Real-world Experimental results. Each method was executed ten times for each scenario. 'S' denotes single-person occlusion, while 'M' indicates multiple-person occlusion.}
    \label{fig:realworld}
\vspace{-13pt}
\end{figure}

\begin{table}[t]
    \centering
    \caption{Person search performance in simulated environments with different scales and room structures involving static and dynamic pedestrians.}
    \scalebox{0.76}{ 
        \begin{tabular}{l|cc|cc|cc}
            \toprule 
            \multirow{2}{*}{\textbf{Method}} &\multicolumn{2}{c|}{\textbf{Factory$^*$}} &\multicolumn{2}{c|}{\textbf{Hospital$^*$}} &\multicolumn{2}{c}{\textbf{Bookstore$^*$}} \\
            & \textit{SR}\textuparrow &\textit{SPL}\textuparrow & \textit{SR}\textuparrow & \textit{SPL}\textuparrow &\textit{SR}\textuparrow &\textit{SPL}\textuparrow \\
            \midrule
            \midrule
            \textbf{GTLL\cite{chen2017integrating}} &34 &32.6 &32 &31.0 &58 &56.0 \\
            \textbf{GTPL\cite{Kim2018AnAF}} &\underline{80} &\underline{73.9} &49 &46.6 &50 &48.8 \\
            \textbf{GTLL\cite{chen2017integrating} + Greedy-NBV} &61 &61.5 &\underline{84} &\underline{81.4} &\underline{88} &81.5 \\
            \textbf{GTPL\cite{Kim2018AnAF} + Greedy-NBV} &73 &67.6 &65 &55.9 &70 &62.9 \\
            \textbf{Active Graph-SLAM\cite{placed2023ral}} &19 &18.4 &49 &48.0 &67 &66.2 \\
            \textbf{HB-Particle\cite{goldhoorn2017search}} &8 &6.9 &7 &4.3 &83 &69.7 \\
            \rowcolor{gray!25}
            \textbf{RPF-Search (ours)} &\bf 100 &\bf 94.7  &\bf 98 &\bf 94.8 &\bf 98 &\bf 95.7 \\
            \bottomrule
        \end{tabular}
    }
    \label{tab:SRandSTinAll}
\end{table}

\begin{figure*}[h]
    \centering
    \includegraphics[width=1.0\textwidth, trim={0pt 0pt 0pt 0pt}, clip]{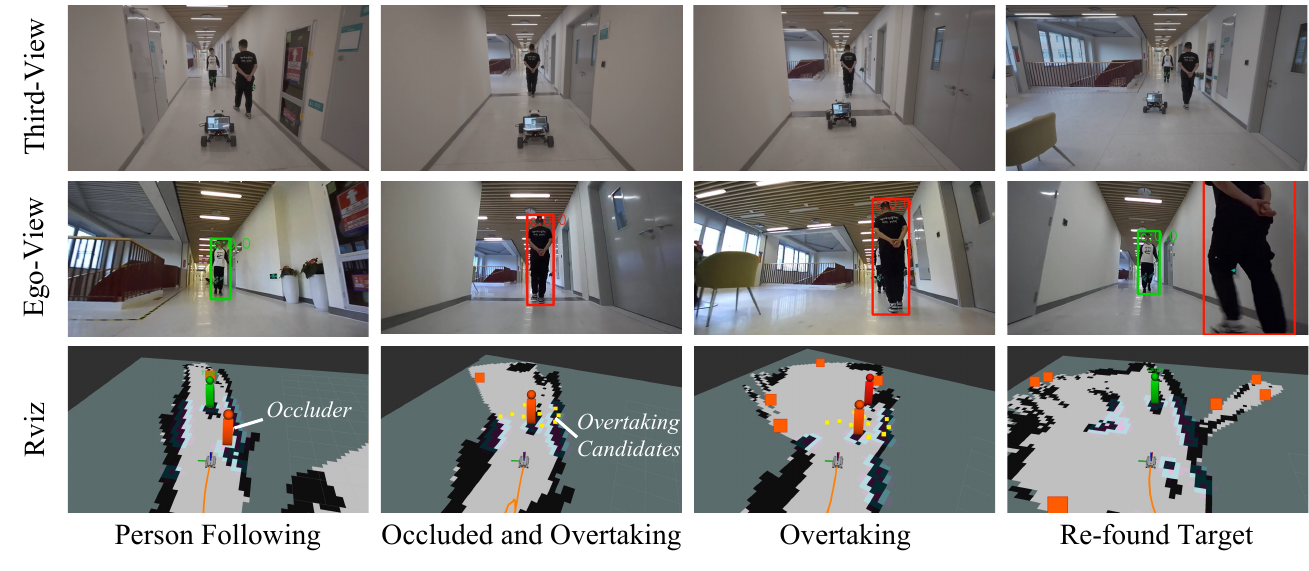}
    \caption{Keyframes from our method during real-world experiments involving single-person occlusion in overtaking scenarios. When the target person is occluded by a pedestrian, an observation-based overtaking field is constructed. Overtaking candidates (represented by yellow points) are sampled and evaluated within this field. The candidate with the minimum cost, provided that the cost is below a predefined threshold, is selected as the next search point. Through continuous selection of search points and robot navigation, the robot smoothly performs the overtaking maneuver and successfully re-finds the target person.}
    \label{fig:RealSingleDynaVis}
\vspace{-10pt}
\end{figure*}

\begin{figure*}[h]
    \centering
    \includegraphics[width=1.0\textwidth, trim={0pt 0pt 0pt 0pt}, clip]{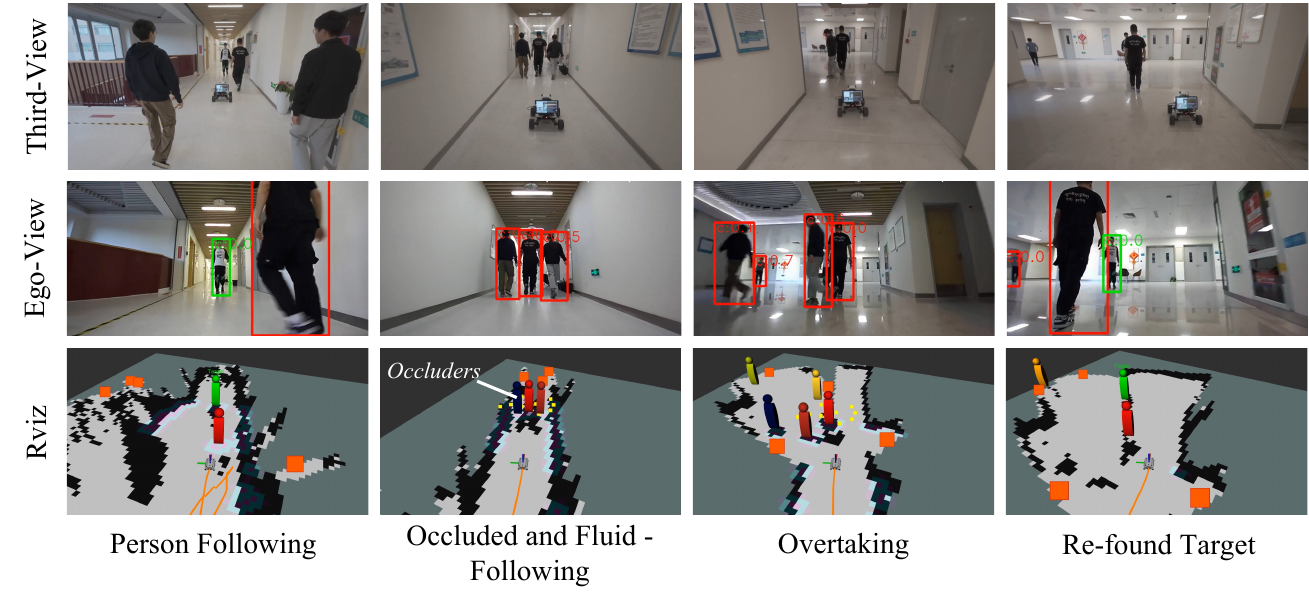}
    \caption{Keyframes from our method during real-world experiments involving multi-person occlusion in a narrow corridor. When the target person is occluded due to dynamic obstructions, an overtaking field is first constructed. However, because the minimum overtaking cost exceeds a predefined threshold, a fluid field based on the motion cues of visible occluders is created to ensure safe navigation and facilitate target searching. The robot follows the fluid field (shown in the second column) until an overtaking opportunity arises (depicted in the third column), enabling the robot to re-find the target person.}
    \label{fig:RealMultiDynaVis}
\vspace{-15pt}
\end{figure*}

\begin{figure}
    \centering
    \includegraphics[width=0.65\linewidth, trim={0pt 0pt 0pt 0pt}, clip]{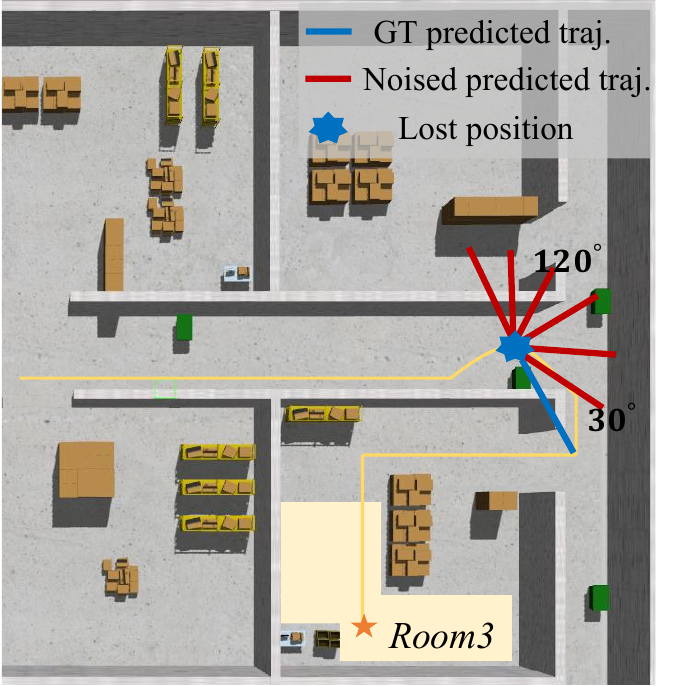}
    \caption{Environmental setting of ablation study about different accuracies of trajectory predictions. We simulate degraded predictor performance by systematically rotating the ground-truth future trajectory orientation within the range of $[0^{\degree},180^{\degree}]$. This range is uniformly discretized into seven distinct trajectories.}
    \label{fig:ablationTraj}
\vspace{-15pt}
\end{figure}

\begin{table}[t]
    \centering
    \caption{Effect of trajectory-prediction accuracy on person-search performance. We simulate different trajectory-prediction accuracies by systematically rotating the ground-truth trajectory. ``Rotation'' denotes the manually applied rotation angle relative to the ground-truth trajectory. A larger value represents a worse prediction performance.}
    \scalebox{0.8}{
    \begin{tabular}{l|ccccccc}
    \toprule 
    \textbf{Rotation} &$+0^{\degree}$ &$+30^{\degree}$ &$+60^{\degree}$ &$+90^{\degree}$ &$+120^{\degree}$ &$+150^{\degree}$ &$+180^{\degree}$ \\
    \midrule
    SR &99 &95 &91 &90 &42 &18 &6 \\
    SPL &94.9 &90.5 &86.1 &85.2 &38.4 &16.3 &5.4\\
    \bottomrule
    \end{tabular}
    }
    \label{tab:ablationTraj}
\vspace{-10pt}
\end{table}

\begin{table}[t]
    \centering
    \caption{Computational cost of the belief-guided search field updating in different-scale scenarios.}
    \scalebox{0.9}{
    \begin{tabular}{l|ccc}
    \toprule 
    \textbf{Scenario} &Bookstore &Hospital &Factory \\
    \midrule
    Time cost (ms) &$14.5\pm 6.5$ &$26.4\pm 9.1$ &$10.1\pm 2.9$  \\
    \bottomrule
    \end{tabular}
    }
    \label{tab:timeCost}
\vspace{-15pt}
\end{table}

The real-world experiments were conducted in an indoor office environment featuring a long hallway and a large U-turn, specifically designed to test the performance of the algorithm under dynamic and topographic occlusions. In the topographic occlusion experiment (Fig. \ref{fig:coverTopographic}, our proposed algorithm demonstrated significant advantages over GTLL\cite{chen2017integrating} + Greedy-NBV and GTPL\cite{Kim2018AnAF} + Greedy-NBV, which were the best-performing algorithms in previous comparisons. Notably, our approach achieved perfect tracking recovery, with ten successful runs out of ten --- eight more than GTLL and seven more than GTPL, as shown in Fig.~\ref{fig:realworld}. This improvement is due to the limitations of GTLL + Greedy-NBV and GTPL + Greedy-NBV, as their reliance on local greedy search often results in suboptimal paths, such as frequent detours or overly conservative trajectories. In contrast, our algorithm leverages a probabilistic distribution to optimize the search process, reducing unnecessary exploration and producing more efficient and natural navigation paths.

In the dynamic occlusion experiments (Figs \ref{fig:coverSingle} and \ref{fig:coverMulti}), our algorithm consistently outperformed the compared methods. Unlike existing approaches, which struggle with dynamic occlusions, our algorithm integrates adaptive strategies—such as fluid-following and overtaking—that effectively address varying occlusion scenarios in real-world environments. These adaptive strategies significantly improve the robot’s ability to re-find the target person after occlusions, resulting in a higher success rate. Specifically, our method achieved 7 more successes in Long Hallway (S) and 8 more successes in Long Hallway (M) compared to the version without adaptive strategies, as shown in the experimental data (Fig.~\ref{fig:realworld}).

Visualizations of the single-person and multi-person occlusion experiments are provided in Fig.~\ref{fig:RealSingleDynaVis} and Fig.~\ref{fig:RealMultiDynaVis}, respectively. In the single-person occlusion scenario, an observation-based overtaking field is constructed using the motion cues of the occluder. Within this field, overtaking candidates (indicated by yellow points in the figure) are sampled and evaluated. The candidate with the minimum cost, as long as it is below a predefined threshold, is selected as the next search point. Through continuous selection of search points and navigation, the robot smoothly overtakes the occluder and successfully re-finds the target person.

In the multi-person occlusion scenario, our method first constructs the overtaking field. However, because the minimum overtaking cost exceeds a predefined threshold, a fluid field is created using the motion cues of visible occluders to ensure safe navigation and facilitate target searching. The robot follows the fluid field (shown in the second column) until an overtaking opportunity arises (shown in the third column), allowing the robot to resume tracking the target.

\begin{figure}[h]
    \centering
        \begin{subfigure}[t]{0.23\textwidth}
            \centering
            \includegraphics[width=\textwidth, trim={30pt 0pt 50pt 50pt}, clip]{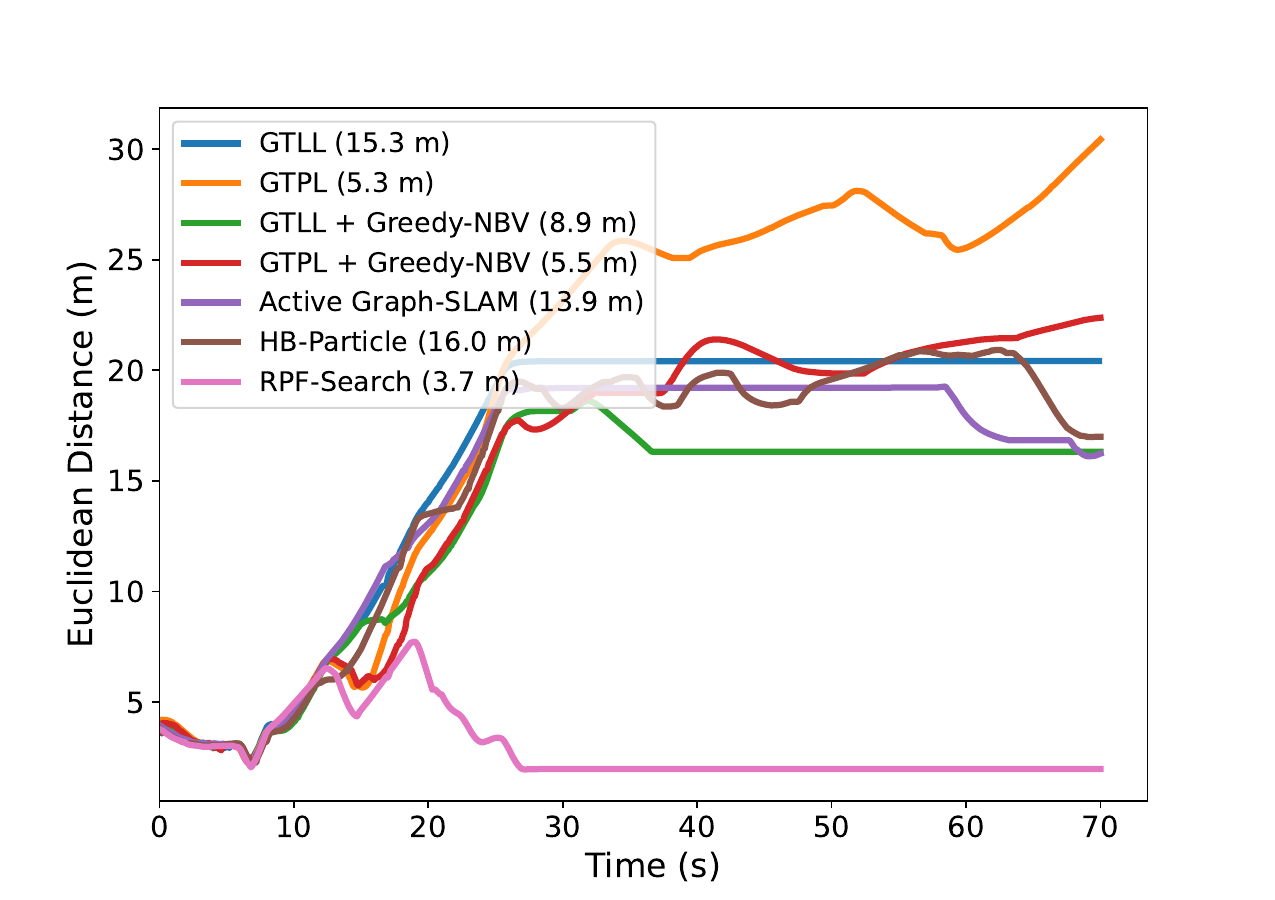}
            \caption{\textit{Room1}}
            \label{fig:topoCompareRoom1}
        \end{subfigure}%
        \hspace{0.001\linewidth}
        \begin{subfigure}[t]{0.23\textwidth}
            \centering
            \includegraphics[width=\textwidth, trim={30pt 0pt 50pt 50pt}, clip]{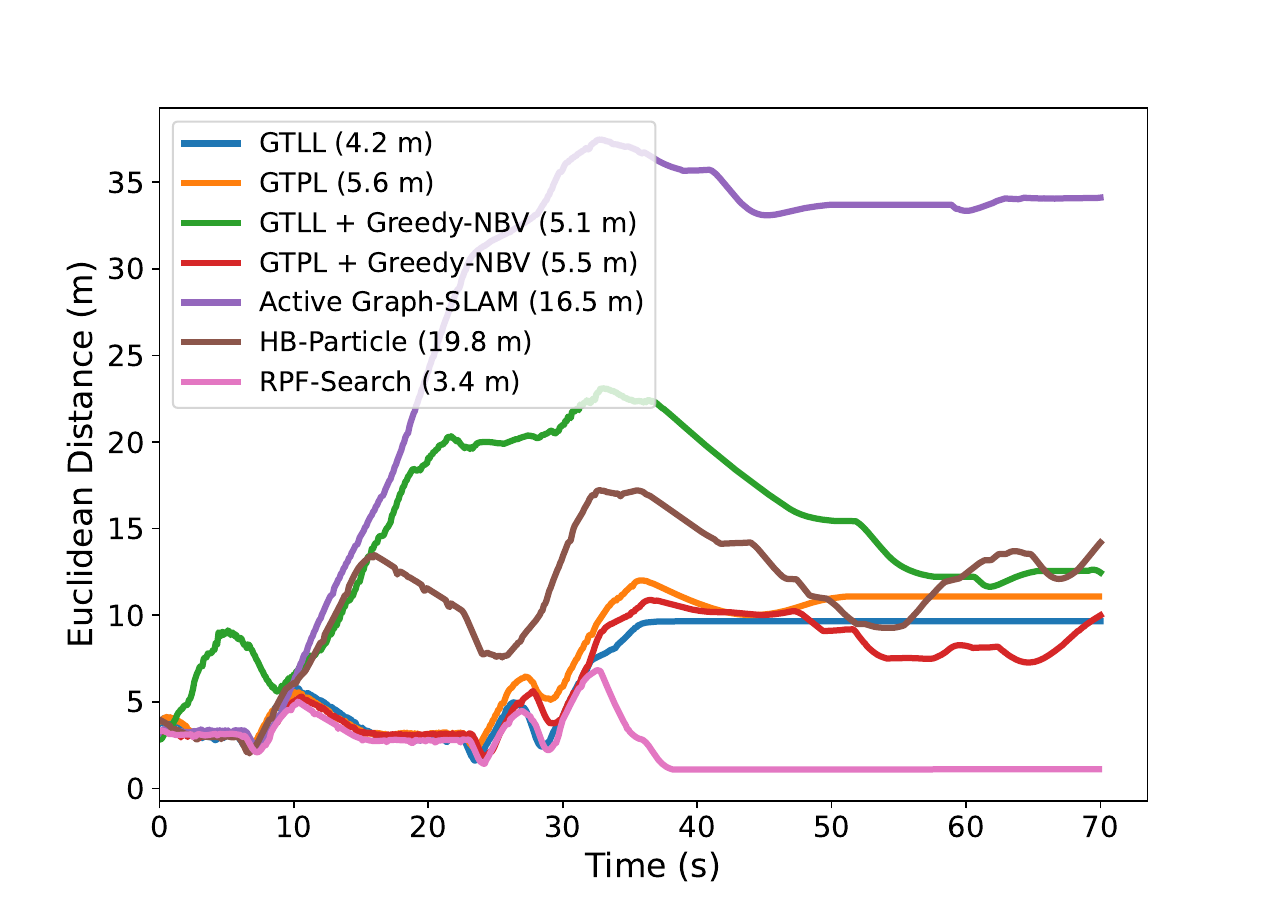}
            \caption{\textit{Room2}}
            \label{fig:topoCompareRoom2}
        \end{subfigure}%
        \\
        \begin{subfigure}[t]{0.23\textwidth}
            \centering
            \includegraphics[width=\textwidth, trim={30pt 0pt 50pt 50pt}, clip]{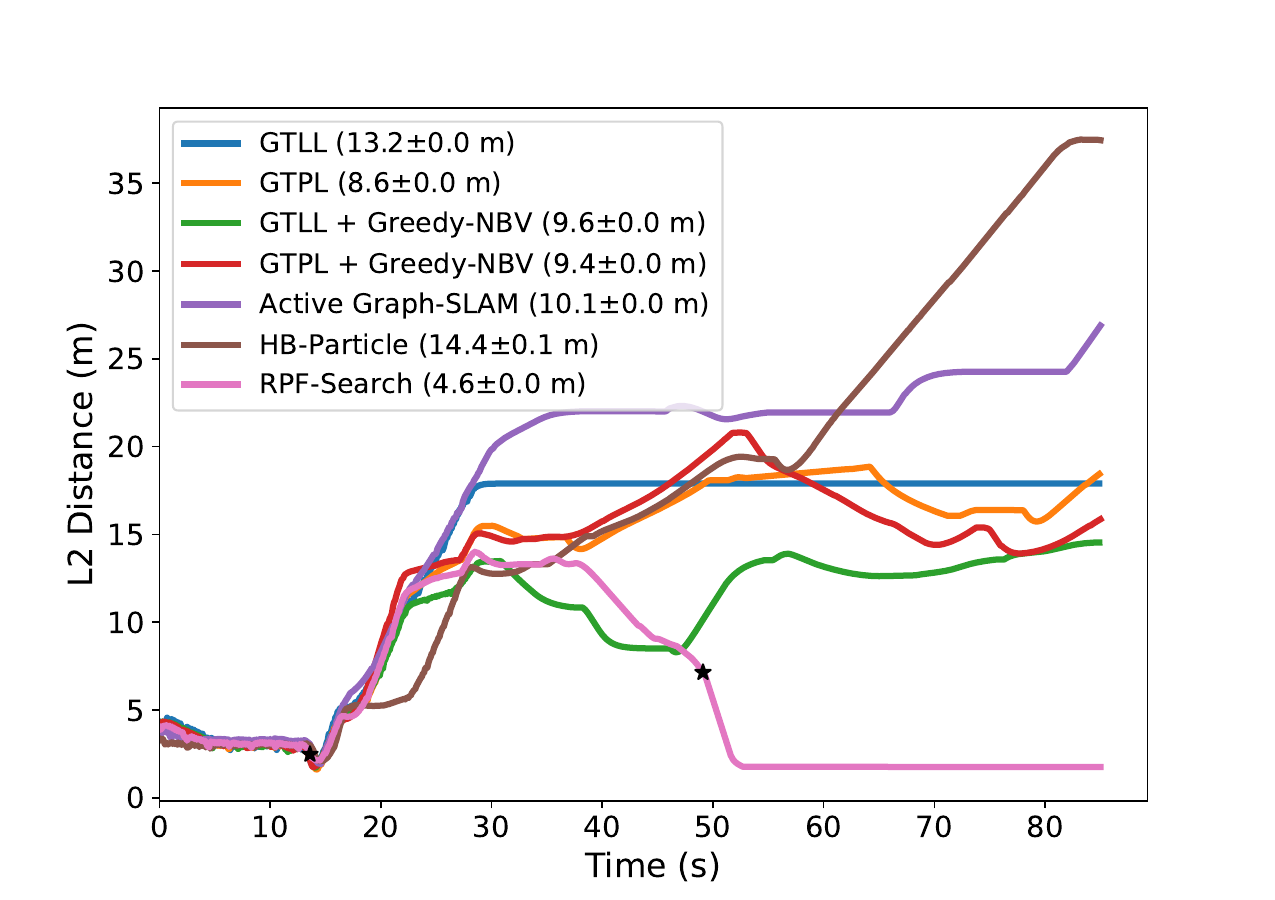}
            \caption{\textit{Room3}}
            \label{fig:topoCompareRoom3}
        \end{subfigure}%
        \hspace{0.001\linewidth}
        \begin{subfigure}[t]{0.23\textwidth}
            \centering
            \includegraphics[width=\textwidth, trim={30pt 0pt 50pt 50pt}, clip]{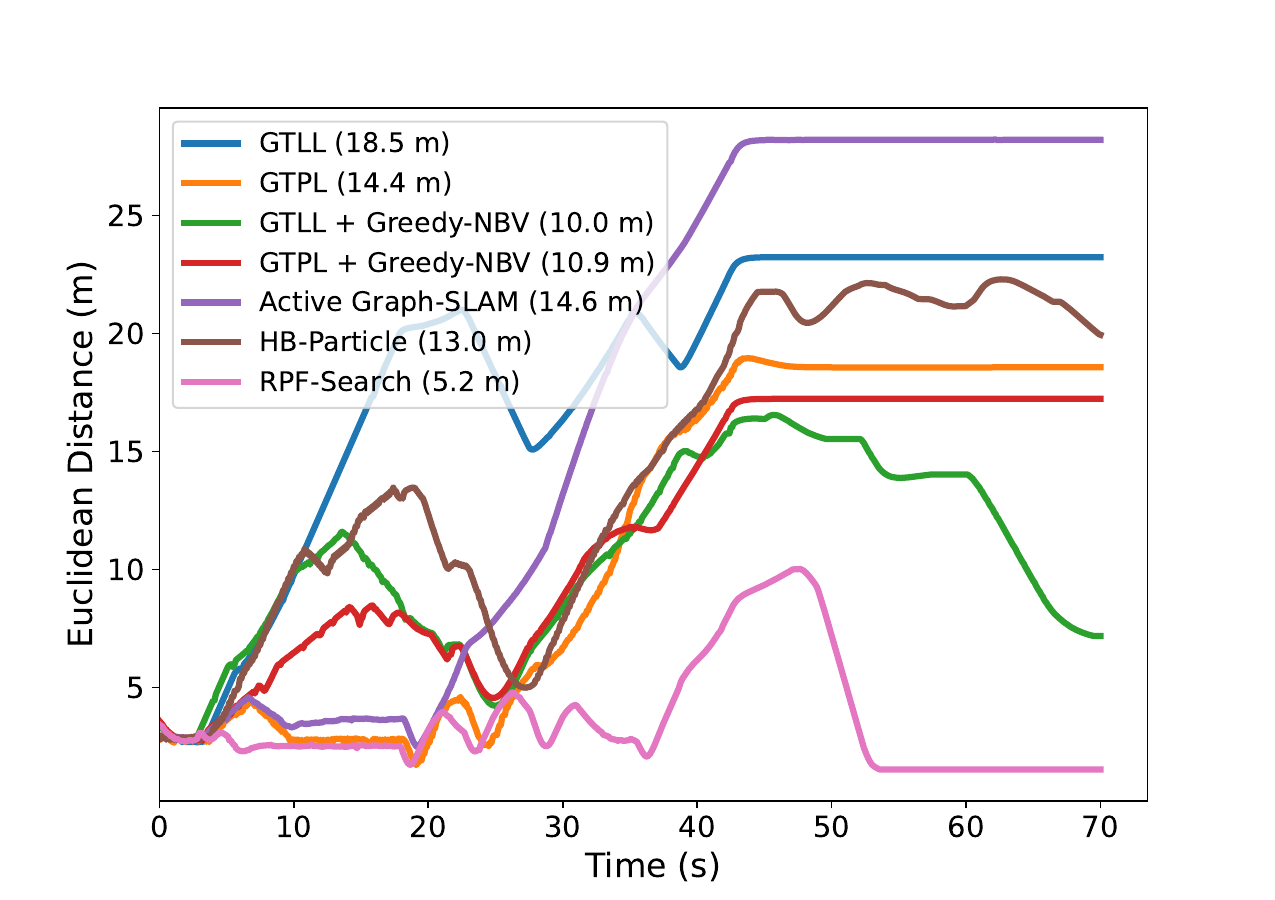}
            \caption{\textit{Dyna1}}
            \label{fig:dynaCompareDyna1}
        \end{subfigure}%
        \\
        \begin{subfigure}[t]{0.23\textwidth}
            \centering
            \includegraphics[width=\textwidth, trim={30pt 0pt 50pt 50pt}, clip]{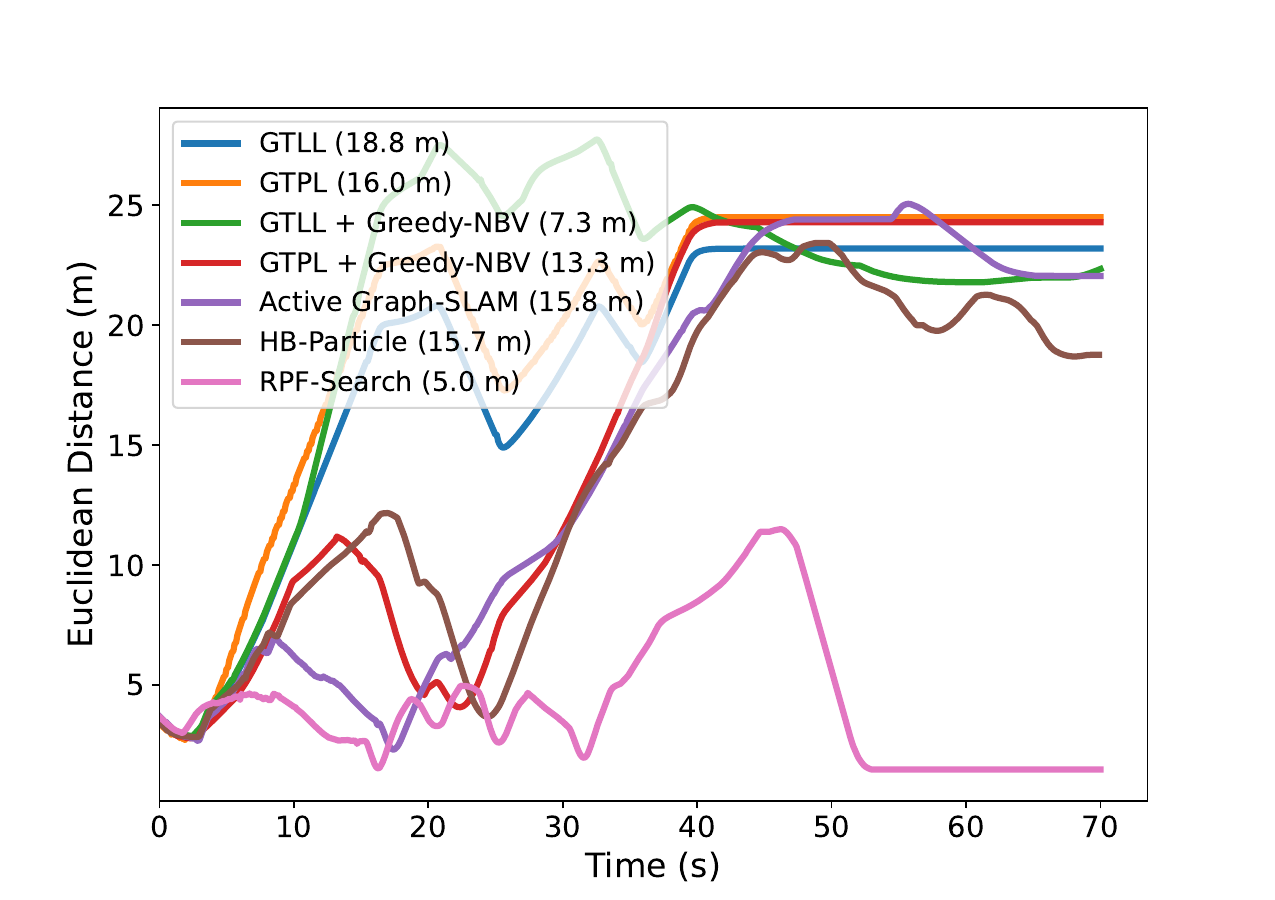}
            \caption{\textit{Dyna2}}
            \label{fig:dynaCompareDyna2}
        \end{subfigure}%
        \hspace{0.001\linewidth}
        \begin{subfigure}[t]{0.23\textwidth}
            \centering
            \includegraphics[width=\textwidth, trim={30pt 0pt 50pt 50pt}, clip]{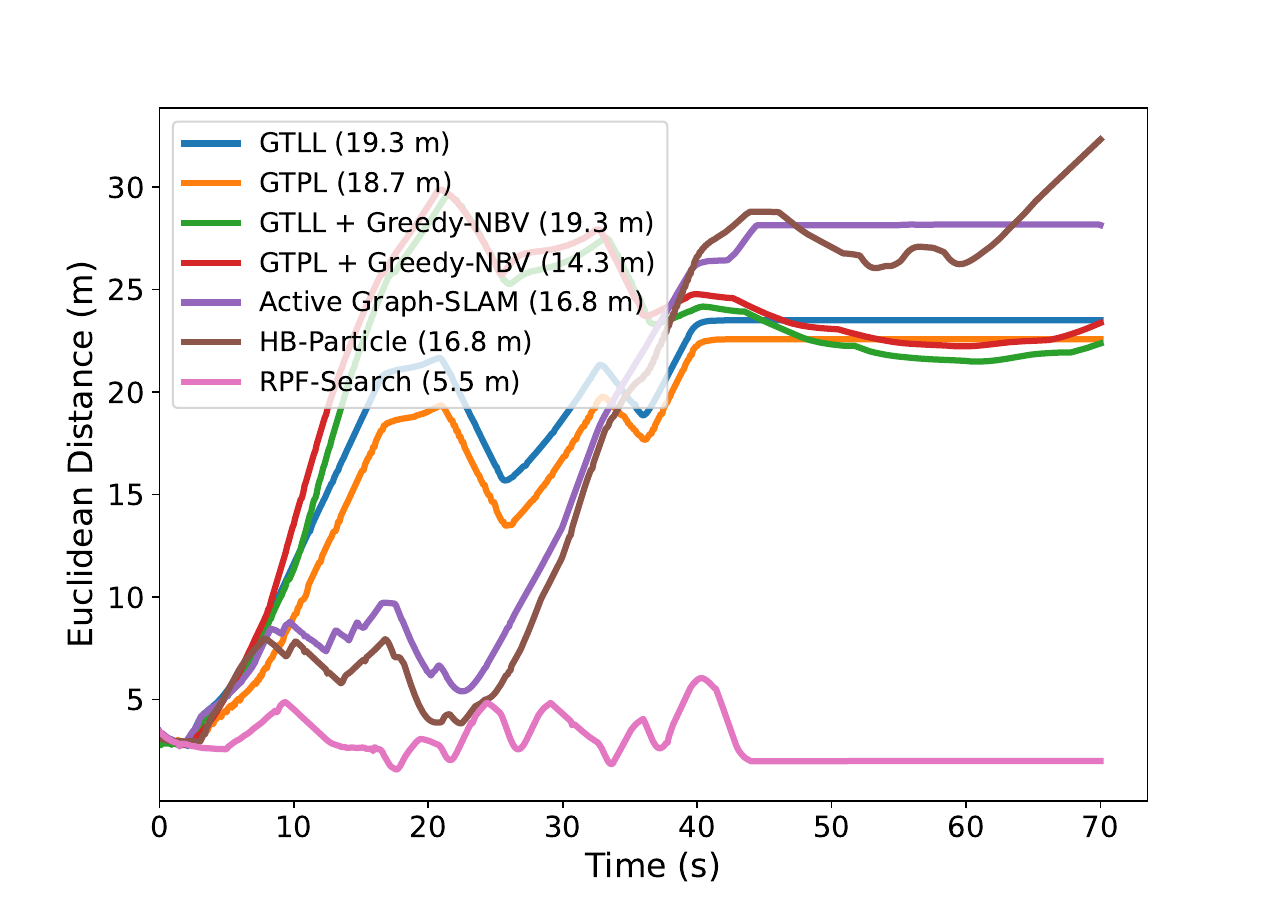}
            \caption{\textit{Dyna3}}
            \label{fig:dynaCompareDyna3}
        \end{subfigure}%
    \centering
    \caption{Euclidean distance between the robot and the target person over time. When the target person was lost, the fluctuations in the distance using our method were smaller than those in other baselines, indicating that our approach could quickly navigate to a true direction to search for the target. Additionally, compared to the baselines, our method converges to the person-following distance more rapidly. The average distance, calculated over 100 runs, is shown in the legend. Our method consistently achieves the best performance, with average distances of 3.7 m in \textit{Room1}, 3.4 m in \textit{Room2}, 6.4 m in \textit{Room3}, 5.2 m in \textit{Dyna1}, 5.0 m in \textit{Dyna2}, and 5.5 m in \textit{Dyna3}.}
    \label{fig:distCompare}
\vspace{-15pt}
\end{figure}

\subsection{Simulated Experiments}
\textbf{Evaluation on different scales and room structures.} The evaluation results of different scales and room structures are shown in Table~\ref{tab:SRandSTinAll}, our RPF-Search achieves the highest performance, with 100\% SR and 94.7\% SPL in \textit{Factory}, 98\% SR and 94.8\% SPL in \textit{Hospital}, and 98\% SR and 95.7\% SPL in \textit{Bookstore}. It can be observed that person search in small-scale environments (e.g., \textit{Bookstore}) is relatively straightforward, with five baselines achieving an SR above 50\%, and the second-best method reaching an SR of 88\% SR and an SPL of 81.5\%. This may be attributed to the fact that the target person is more easily observed as the robot explores the smaller Bookstore environment. In contrast, person search becomes more challenging in larger environments (e.g., \textit{Hospital} and \textit{Factory}), where incorrect decisions can lead the robot to move away from the target person. Consequently, only two baselines achieve above 50\% SR in \textit{Hospital}, and three achieve above 50\% SR in \textit{Factory}.

Our method, however, consistently outperforms the second-best approach, with improvements of +20\% in terms of SR and +20.8\% in terms of SPL in \textit{Factory}, +14\% SR and +13.4\% SPL in \textit{Hospital}, and +10\% SR and +14.2\% SPL in \textit{Bookstore}. This demonstrates the effectiveness of our belief-guided field, which leverages newly observed environmental information and the target person's past motion cues to update the belief of the target person's location. This approach guides the robot to explore the most probable areas, ultimately leading to a successful re-identification of the target person.

\textbf{Effect of trajectory prediction accuracy.} To investigate how the accuracy of trajectory prediction affects the person-search performance, we conducted an ablation study. This study focuses on the \textit{Room3} scenario, which presents the most challenging case due to its complex intersection layout and the target's distinctive U-turn trajectory pattern. To evaluate robustness, we simulate degraded predictor performance by systematically rotating the ground-truth future trajectory orientation within the range of $[0^{\degree},180^{\degree}]$. This range is uniformly discretized into seven distinct trajectories (as shown in Fig.~\ref{fig:ablationTraj}), each representing a different angular deviation. Using these perturbed trajectories, we propagate the target person's existence probability, leading to varying spatial belief distributions in the regions traversed by the target. As a result, the person-search SR and SPL are affected accordingly as shown in Table~\ref{tab:ablationTraj}. We observe a significant degradation in person-search performance once the predicted trajectory is rotated by about $+120^{\degree}$, with SR dropping to 42\% and SPL to 38.4\%. For rotation angles below $+90^{\degree}$, both SR and SPL remain above 90\% and 85.2\%, respectively. This suggests that our belief-guided search field is resilient to noisy trajectory predictions, owing to its probabilistic design.

\textbf{Computational cost of the belief-guided search field updating in different-scale scenarios.} To assess the real-time performance of the proposed belief-guided search field updating method across scenarios of varying scales, we conduct a comprehensive computational cost analysis. The results, presented in Table~\ref{tab:timeCost}, demonstrate that our method achieves the lowest computational cost of $10.1 \pm 2.9$ ms in the factory scenario—despite it being the largest-scale environment ($40\times70$ $m^2$). In contrast, the mid-scale hospital scenario ($26\times50$ $m^2$) requires $26.4 \pm 9.1$ ms, which is still highly efficient.

This difference arises because the belief-guided field is constructed based on frontier detections. Cluttered environments, such as the hospital, introduce more occlusions, leading to increased frontier detections and thus higher computational costs. Although the factory is larger in scale, its lower object density reduces occlusion complexity, resulting in faster computation. Notably, all observed computational costs remain at a practical level, with the highest being only $26.4 \pm 9.1$ ms. These results confirm that the proposed belief-guided field enables real-time decision-making for robotic systems across diverse environments.

\textbf{Euclidean distance to the target over time.} For topographic occlusion, as observed in Fig.~\ref{fig:topoCompareRoom1}, Fig.~\ref{fig:topoCompareRoom2} and Fig.~\ref{fig:topoCompareRoom3}, when using our RPF-Search method, the distance between the robot and the person quickly converges to the desired following distance after tracking loss. Consequently, as indicated in the legends of these figures, our method achieves the smallest average Euclidean distance to the target, with 3.7 m in \textit{Room1}, 3.4 m in \textit{Room2} and 6.4 m in \textit{Room3}, surpassing the second-best baseline by reductions of 1.8 m, 0.8 m and 3.0 m, respectively. This demonstrates that our method can help the robot quickly re-find the target person and return to the person-following stage.

For dynamic occlusion, EDTOT results shown in Fig.~\ref{fig:dynaCompareDyna1}, Fig.~\ref{fig:dynaCompareDyna2} and Fig.~\ref{fig:dynaCompareDyna3} also evidence the effectiveness of our method compared to other baselines. Our method can achieve the lowest average distance with 5.2 m in \textit{Dyna1} and 5.0 m in \textit{Dyna2} and 5.5 m in \textit{Dyna3}, surpassing the second-best baseline by reductions of 4.8 m, 2.3 m and 8.8 m, respectively. These significant improvements confirm the effectiveness of our RPF-Search method in handling dynamic occlusions. By incorporating our proposed fluid-following and overtaking fields, the robot can perform safe and smooth search behaviors when dealing with dynamic occlusions.

\begin{figure*}[h]
    \centering
    \includegraphics[width=1.0\textwidth, trim={8pt 0pt 0pt 0pt}, clip]{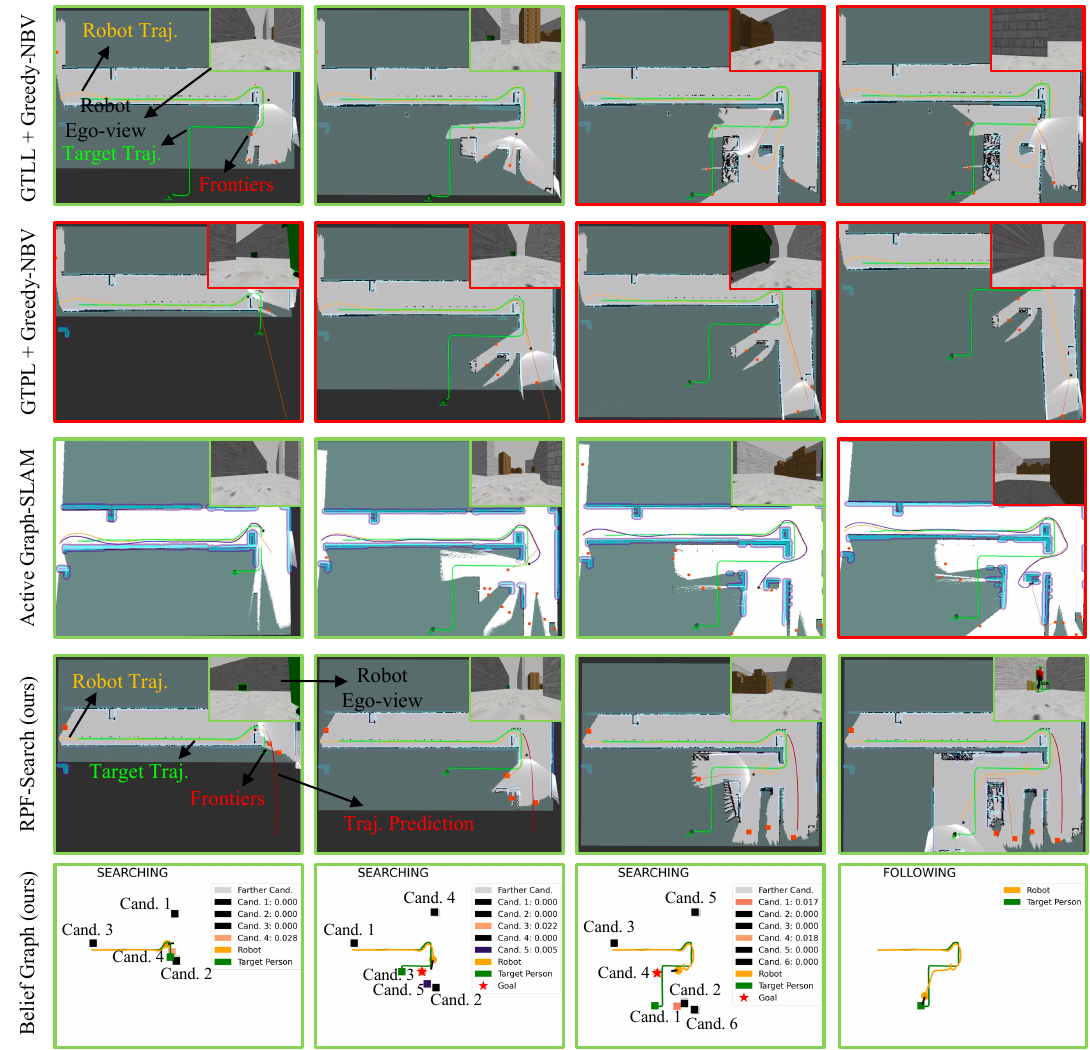}
    \caption{Searching keyframes of baselines and our method in topographic occlusion scenarios in \textit{Room3}. The first row shows the explored occupancy map, frontiers (red squares), robot and target positions, ego view, and trajectories (orange and green lines). The second row presents the belief-guided graph, including robot and target trajectories, candidate frontiers with belief scores (e.g., “Cand. 4: 0.028”), and additional frontiers for probabilistic inheritance (grey squares). Initially, trajectory predictions direct the robot toward the corridor (red line), but our belief-guided graph updating integrates new environmental data and the target’s motion history to penalize unlikely corridor candidates and focus the search on more promising areas within the room.}
    \label{fig:topoVisRoom3}
\end{figure*}

\textbf{Visualization of baseline performance in \textit{Room3}} Moreover, we provide a person-search visualization of baselines and our method in \textit{Room3} as shown in Fig.~\ref{fig:topoVisRoom3} where the target person's route involves making a U-turn corner and entering a room. The visualizations of our method are provided in the last two rows of Fig.~\ref{fig:topoVisRoom3}. Initially, the trajectory prediction (depicted as a red line in the fourth row) directs the robot toward the corridor. However, our method effectively utilizes newly observed environmental information to penalize infeasible candidates through the inference factor (Eq.~\ref{eq:inferenceFactor}). This factor dictates that if the target person were indeed moving toward the corridor, the robot should have observed them by now. Consequently, as shown in the second belief-guided graph, the belief associated with the corridor candidate reduces to zero; specifically, the belief of 'Cand. 2' is zero in the second graph. This prompts the robot to redirect its search toward higher-belief candidates within the room, i.e., 'Cand. 3,' which holds a belief value of 0.022 in the second graph. Without this inference factor, the robot would pursue the target person into the corridor, following the trajectory prediction and ultimately resulting in a failed search.

Additionally, the probabilistic inheritance capability of the belief-guided field (Eq.~\ref{eq:probabilityInheritance}) allows newly observed candidates to inherit belief from parents in a belief-decaying manner. This mechanism enables the robot to continuously adjust its focus toward more probable candidates until locating the target, as shown in the third and fourth graphs of Fig.~\ref{fig:topoVisRoom3}. Due to probabilistic inheritance, as seen in the third graph, 'Cand. 4' inherits belief from its predecessor in a decayed form, resulting in a belief of 0.018, which guides the robot to continue searching the room until the target person is found. If probabilistic inheritance were removed, the belief of 'Cand. 4' would drop to zero due to its distance from the probability area defined by the trajectory prediction, causing the robot to navigate incorrectly and leading to a failed search. Our proposed probabilistic inheritance utilizes the local connectivity of newly produced and old frontiers, enabling an optimal selection of inside-room and outside-room explorations. 

\subsection{Limitations and Discussion}
\textbf{Strategy switch problem.} Our method demonstrates promising performance for person search in unknown and dynamic environments, but it is not yet an optimal solution. For instance, the current approach relies on a naive Intersection over Union (IoU)-based assessment to trigger the two fields. However, in more complex environments where the target person might transition from dynamic occlusion to topographic occlusion, this simplistic approach can mislead the robot. To address this, a more sophisticated occlusion identification method could be explored. One potential solution is leveraging a large language model or a visual-language model to engage in chain-of-thought reasoning, enabling the system to infer the target person's occlusion state by analyzing occluders' motion cues and the surrounding environment.

\textbf{Limitations and potential improvements in handling topographic occlusion.} Furthermore, our method faces challenges in handling long-term disappearances caused by topographic occlusion. Currently, the exclusion of impossible candidates relies heavily on an inference factor that is strongly tied to the duration of the disappearance. When the target is lost for an extended period, this factor becomes uniform across all candidates (frontiers), forcing the method to depend solely on trajectory prediction and probability inheritance. This limitation can result in prolonged search times. To mitigate this, incorporating semantic cues could improve the exclusion process. For example, if prior knowledge indicates that the target person is looking for a drink with an empty bottle, locations such as meeting rooms or gyms would be less likely to contain the person, allowing for more focused and efficient searching.

Additionally, our method can be seamlessly integrated with map-dependent person-search approaches. By dynamically constructing occupancy maps and frontier-based topological graphs during the RPF process, it naturally generates a topo-metric map. The frontiers, serving as nodes, represent safe navigation points and provide maximal observational value. Moreover, during RPF, if the robot accompanies a single user in the same location for an extended period, these nodes can encode the user's appearance frequency, enabling a more efficient and targeted person search.

In the experiments of person search in large-scale environments, we did not observe noticeable delays in map updates, even in our largest test environment—a factory floor with multiple rooms and hallways. This efficiency is primarily due to the fact that active mapping is constrained to the robot’s sensor range; the system does not attempt to update the entire global map at high frequency, but rather focuses on the local vicinity. This inherently bounds the computational load.

Another concern for searching in large-scale environments is the risk of drift or latency during extended operations. To address this problem, we plan to incorporate periodic SLAM-based pose corrections to prevent localization drift over long distances, and adopt a sliding map window strategy—maintaining a fixed-size local map around the robot for updates, while pruning or downsampling distant regions. These strategies help keep update costs consistent and minimize drift, even in expansive environments.

Moreover, potential improvements can be derived from advanced trajectory prediction. For example, by integrating multi-modal goal estimation methods, we can enhance the goal location prediction module in our approach. Specifically, BiTraP\cite{yao2021bitrap} can be employed to predict multiple plausible target locations, and these multi-modal predictions can then be incorporated into our current belief-field construction, significantly improving robustness and success rates by better accounting for uncertainties in target movements. Improving local environmental perception is another promising option. Integrating a Conv-LSTM-based module\cite{song2020pedestrian} into our framework will more effectively capture spatial interactions between pedestrians and the environment, enhancing the accuracy of environmental predictions in our probabilistic search field and thus boosting search efficiency and adaptability.

\textbf{Limitations and potential improvements in handling dynamic occlusion.} In our person-search task, dynamic obstacles predominantly consist of pedestrians. To enhance robot-pedestrian coexistence, we implement socially conscious navigation strategies rooted in HRI principles. Our robot’s navigation decisions are influenced by the motions of people in its vicinity via the observation-based potential field (Sec.~\ref{sec:DynaSearch}). For instance, if a pedestrian (not the target) is moving and causing a dynamic occlusion, our robot will either follow behind that person or choose to overtake them, depending on the situation; this is effectively the robot reacting to human behavior in real time. We have highlighted that this fluid-following approach provides a degree of implicit social compliance – the robot doesn’t aggressively push through crowds but rather moves with the flow, which is a desirable HRI trait.

In addition, explicit social reasoning is crucial for improving human-robot interaction. In future work, we plan to integrate human intent inference by fusing gaze and body pose cues with our SVR-based predictor, enabling the robot to anticipate sudden stops or evasive movements. We also intend to incorporate semantic affordance detection—such as recognizing doorways, elevator entrances, and similar transitions—to propagate belief across such semantic portals, as suggested by recent advances in semantic HRI. These modules are algorithmically compatible with our existing framework and can be incorporated as additional weights or constraints without disrupting the real-time pipeline.

In highly dynamic environments with fast or dense human traffic, sensor occlusion and inconsistent observations can still lead to residual artifacts or uncertain regions. To mitigate this, a promising extension is to implement dynamic object filtering: leveraging computer vision or LiDAR clustering to explicitly detect and exclude dynamic entities such as humans from contributing to the occupancy grid update. In this approach, dynamic obstacles are tracked separately, preventing them from corrupting the static map. While our current implementation does not yet include such filtering, we consider it a practical and effective addition for improving mapping robustness. In addition, our fluid-following mechanism inherently reduces reliance on a perfectly accurate map when dealing with dynamic obstacles. Since the robot responds to moving humans via a real-time potential field, it can adaptively flow around them—even if they are not explicitly mapped as obstacles.


\end{document}